\documentclass{article}

\usepackage[final]{neurips_2025}

\usepackage[utf8]{inputenc} 
\usepackage[T1]{fontenc}    
\usepackage{hyperref}       
\usepackage{url}            
\usepackage{booktabs}       
\usepackage{amsfonts}       
\usepackage{nicefrac}       
\usepackage{microtype}      
\usepackage{xcolor}         
\usepackage{wrapfig}
\usepackage{conference}
\usepackage[whole]{bxcjkjatype}

\DeclareMathOperator*{\argmin}{argmin}

\usepackage{booktabs, multirow, colortbl, xcolor, pifont}
\arrayrulecolor{black!40}
\definecolor{linkblue}{RGB}{0,0,153}
\hypersetup{
  colorlinks=true,
  linkcolor=linkblue,
  citecolor=linkblue,
  urlcolor=linkblue,
  pdfborder={0 0 0},
}
\definecolor{qepgray}{RGB}{235,242,255}   
\newcommand{\q}[1]{\cellcolor{qepgray}#1}
\usepackage{fontawesome5}
\newcommand{\codelink}{%
  \faGithub\ \textbf{Code:} \textcolor{blue}{\url{https://github.com/FujitsuResearch/qep}}%
}

\title{Quantization Error Propagation: \\
Revisiting Layer-Wise Post-Training Quantization}

\author{Yamato Arai 
\\
Fujitsu Limited \\  
Department of Basic Science\\
The University of Tokyo\\
\And 
Yuma Ichikawa 
\\
Fujitsu Limited \\
RIKEN center for AIP
\AND
\codelink
}

\begin{document}

\maketitle 
\vspace{-1.8em}
\begin{abstract}
    Layer-wise PTQ is a promising technique for compressing large language models (LLMs), due to its simplicity and effectiveness without requiring retraining. However, recent progress in this area is saturating, underscoring the need to revisit its core limitations and explore further improvements. We address this challenge by identifying a key limitation of existing layer-wise PTQ methods: the growth of quantization errors across layers significantly degrades performance, particularly in low-bit regimes. To address this fundamental issue, we propose Quantization Error Propagation (QEP), a general, lightweight, and scalable framework that enhances layer-wise PTQ by explicitly propagating quantization errors and compensating for accumulated errors. QEP also offers a tunable propagation mechanism that prevents overfitting and controls computational overhead, enabling the framework to adapt to various architectures and resource budgets. Extensive experiments on several LLMs demonstrate that QEP-enhanced layer-wise PTQ achieves substantially higher accuracy than existing methods. Notably, the gains are most pronounced in the extremely low-bit quantization regime.
\end{abstract}

\section{Introduction}\label{sec:intro}
Large Language Models (LLMs) have achieved impressive performance in various natural language processing tasks, including open-ended text generation, multi-step reasoning, and dialogue modeling.
Notable examples include ChatGPT \citep{achiam2023gpt} and the Llama family \citep{touvron2023llama, grattafiori2024llama}. 
However, deploying LLMs cost-effectively remains difficult because of their substantial memory usage and computational demands \citep{chen2023frugalgpt}.
This limitation is especially critical for edge computing and latency-sensitive applications. 
To address these challenges, a wide range of model compression techniques, such as quantization~\citep{lang2024comprehensive, gong2024survey}, pruning~\citep{wang2024model, cheng2024survey}, low-rank approximation~\citep{yang2024low, hu2022lora}, and knowledge distillation~\citep{xu2024survey, yang2024survey}, have been explored.

Among these methods, layer-wise post-training quantization (PTQ) has emerged as a practical and widely used approach for large-scale LLMs \citep{frantar2022gptq, lin2024awq, yao2022zeroquant, chee2023quip}. 
Unlike block-wise PTQ \citep{tseng2024quip, shao2023omniquant}, global fine-tuning \citep{egiazarian2024extreme,tseng2024quip}, quantization-aware training (QAT)~\citep{xu2024onebit, wang2023bitnet, liu2023llm}, and all of which require heavy retraining and backpropagation, 
layer-wise PTQ quantizes model parameters layer-by-layer without retraining or backpropagation, resulting in significantly lower computational and memory demands.
Despite its simplicity, layer-wise PTQ effectively preserves model quality even at lower bit widths \citep{frantar2022gptq, lin2024awq, chee2023quip}.
As a result, layer-wise PTQ is increasingly adopted in real-world applications due to its efficient quantization, reduced computational cost, and broader compatibility with large-scale LLMs, varying bit widths, and diverse quantization strategies. 

Despite significant progress in layer-wise PTQ, 
advancements in this area are saturating \citep{malinovskii2024pv}.
This study aims to push the performance boundaries of layer-wise PTQ by revisiting its core design strategy.
This study begins by identifying a fundamental limitation of existing layer-wise PTQ approaches. 
These approaches do not adequately account for the propagation of quantization errors across layers.
Quantization errors accumulate significantly, leading to a degradation in overall model performance, especially in low-bit settings.
This represents a key bottleneck for the practical deployment of layer-wise PTQ in large-scale LLMs.

\begin{figure}
    \centering
    \label{fig:main-results}
    \includegraphics[width=1.0\linewidth]{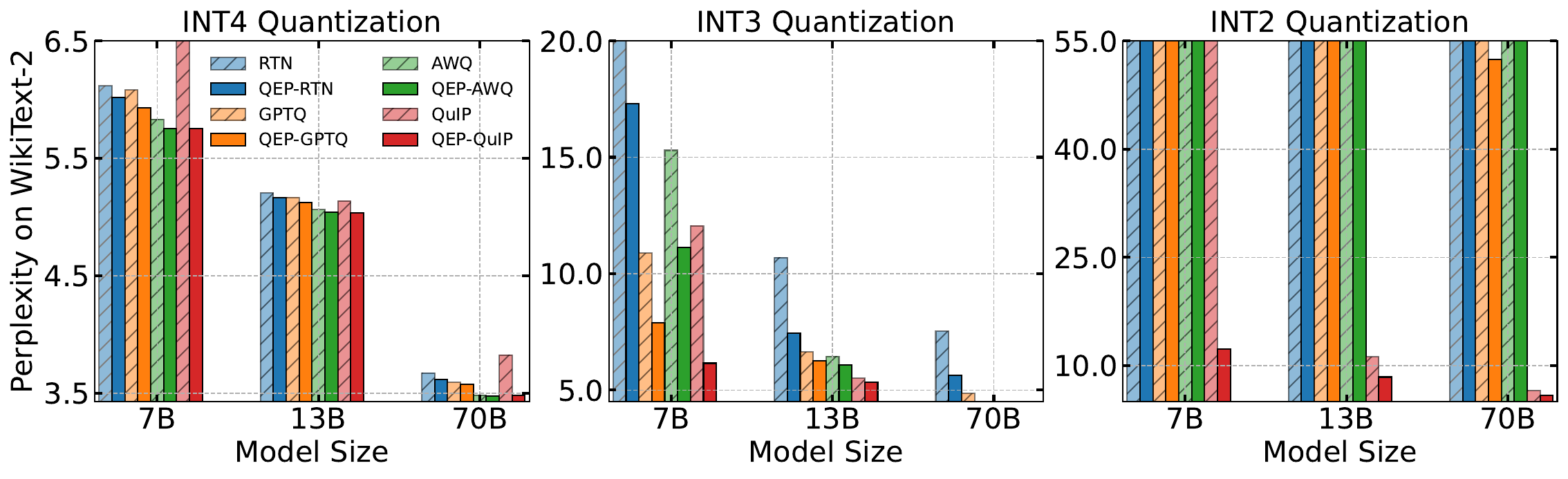}
    \caption{WikiText‑2 perplexity comparison across Llama‑2 models (7B-70B) quantized to INT‑4, INT‑3, and INT‑2, employing RTN, GPTQ, AWQ, and QuIP methods. 
    Solid bars indicate PTQ with QEP; border bars represent PTQ without QEP. Truncated bars indicate perplexities exceeding axis limits. QEP consistently reduces perplexity, with greater improvements observed at lower bitwidths and smaller model sizes. See Section \ref{sec:experiments} for detailed settings and results.}
    \vspace{-10pt}
\end{figure}

To address this issue, we propose \textbf{Q}uantization \textbf{E}rror \textbf{P}ropagation (\textbf{QEP}), a general and computationally efficient framework that enhances the performance of layer-wise PTQ methods. 
QEP modifies the layer-wise optimization objective to propagate and compensate for accumulated quantization errors, while maintaining computational complexity comparable to existing layer-wise PTQ methods.
Furthermore, we introduce a tunable propagation mechanism whose adjustable propagation strength prevents overfitting, a known issue previously observed in GPTQ~\citep{lin2024awq}.
This mechanism also enables adaptive control over computational overhead, especially in parameter-heavy components such as MLP blocks.
Notably, the enhancement of QEP is orthogonal to existing PTQ methods and can be seamlessly integrated with any layer-wise PTQ pipeline.

Extensive experiments on several LLMs across various bit-width settings show that QEP significantly enhances layer-wise PTQ methods, including GPTQ \citep{frantar2022gptq}, AWQ \citep{lin2024awq}, QuIP \citep{chee2023quip}, as shown in Figure \ref{fig:main-results}.
These improvements are particularly pronounced in extreme low-bit regimes, such as 2-bit quantization, where standard layer-wise PTQ methods typically degrade significantly.

\section{Related Work}\label{sec:related-work}
Quantization techniques primarily include data-free PTQ \citep{dettmers2023case}, layer-wise PTQ \citep{frantar2022gptq, lin2024awq, chee2023quip}, block-wise PTQ \citep{tseng2024quip, shao2023omniquant}, global fine-tuning PTQ \citep{egiazarian2024extreme, tseng2024quip}, and QAT \citep{xu2024onebit, wang2023bitnet, liu2023llm}. 
Among these methods, weight-only layer-wise PTQ has become especially popular for large-scale models because of its computational efficiency and strong performance \citep{frantar2022gptq, lin2024awq, chee2023quip}. 
Recent benchmarking further highlights that most PTQ advances specifically target layer-wise methods \citep{zhao2025benchmarking}. 
Following the taxonomy in \citep{zhao2025benchmarking}, we outline three distinct approaches and recent developments.

\paragraph{Compensation-based layer-wise PTQ}
This category, pioneered by GPTQ \citep{frantar2022gptq}, uses a sequential quantization strategy, in which model weights are quantized based on the Hessian computed from a calibration dataset, while compensating for subsequent unquantized weights.
Several studies refined the compensation mechanism by improving update rules \citep{behdin2023quantease}, integrating nonlinear quantization schemes \citep{liu2024vptq},  employing adaptive grid selection \citep{zhang2024leanquant}, and using block-wise optimization \citep{guan2024aptq}.

\paragraph{Rotation-based layer-wise PTQ}
A second promising direction, advanced by QuIP \citep{chee2023quip}, involves preprocessing weights through structured rotation matrices to more uniformly redistribute weight magnitudes. 
This approach was improved by randomized Hadamard transforms and block-wise and global fine-tuning optimization \citep{tseng2024quip}. 
Learning-based methods to determine rotation matrices have also been introduced \citep{liu2024spinquant}. 
This rotation-based strategy has also been extended to activation quantization \citep{ashkboos2024quarot}.

\paragraph{Salience-based layer-wise PTQ}
Other approaches focus on identifying and preserving \emph{salient weights}, often using mixed-precision quantization frameworks \citep{dettmers2022gpt3, dettmers2023spqr, shang2023pb}. 
Although mixed-precision methods usually add complexity due to various data types, AWQ \citep{lin2024awq} mitigates these implementation difficulties. AWQ strategically employs a global scaling mechanism to align salient weights with the quantization grid better, simplifying deployment while maintaining high accuracy.

Recent advances in layer-wise PTQ have mainly focused on nonlinear quantization and block-wise and global fine-tuning extensions; however, the fundamental layer-wise optimization has remained largely unchanged since GPTQ \citep{frantar2022gptq}. 
This study revisits this foundational strategy, identifies its key limitations, and proposes improvements, demonstrating performance gains on the fundamental benchmarks such as  GPTQ \citep{frantar2022gptq}, QuIP \citep{chee2023quip}, and AWQ \citep{lin2024awq}. 
Therefore, our contributions complement and are orthogonal to recent advancements, such as nonlinear quantization and structured extensions.

\section{Background}\label{sec:background}
\paragraph{Post-training quantization}
Post-training quantization (PTQ) is a technique that converts the parameters of pre-trained models into discrete quantized representations. 
Formally, let $\B{W}_{l} \in \mab{R}^{n_{l} \times d_{l}}$ denote the pre-trained weight matrix associated with the $l$-th linear operation.
Note that the index $l$ specifically refers to individual linear transformations rather than entire transformer blocks.
The objective of PTQ is to find a quantized approximation $\widehat{\B{W}}_{l} \in \mab{Q}^{n_{l} \times d_{l}}$ that closely approximates the behavior of the original model, preserving performance while reducing computational costs and memory usage.
The set $\mab{Q} \subset \mab{R}$ denotes the discrete quantization domain, which is represented as a finite set of $2^{b}$ distinct quantization levels, referred to as a $b$-bit quantization scheme.
To achieve accurate quantization, many approaches leverage a small calibration dataset. 
Specifically, given a calibration dataset $\B{X} \in \mab{R}^{d_{1} \times m}$ consisting of $m$ samples, these methods aim to find optimal quantized parameters $\widehat{\B{W}}_{l}$ that minimizes the deviation from the performance of the original model.

\paragraph{Layer-wise PTQ}
Layer-wise PTQ has emerged as a promising framework \citep{frantar2022gptq, frantar2022optimal} for compressing large-scale LLMs.
Recent advancements in this area have significantly reduced the computational overhead and memory requirements of deploying LLMs.
Despite methodological differences, existing layer-wise PTQ approaches typically follow a shared sequential quantization scheme, processing each layer independently and sequentially from the input layer toward the output layer.

Formally, these techniques quantize the model parameters $\{\B{W}_{l}\}_{l=1}^{L}$ by solving the following layer-wise \emph{independent} optimization problem:
\begin{equation}
    \label{eq:original-layerwise-general}
    \min_{\widehat{\B{W}}_{l} \in \mab{Q}^{n_{l} \times d_{l}}} \left\| \B{W}_{l} \mathsf{X}_{l} - \widehat{\B{W}}_l \mathsf{X}_{l} \right\|_{F}^{2},
\end{equation}
where $\mathsf{X}_{l}$ denotes the input activations to the $l$-th layer.
This quantization proceeds sequentially from $l = 1$ toward the output layers.
Due to the quadratic form of the reconstruction objective, the associated Hessian, $\mathsf{H}_{l} \coloneqq \mathsf{X}_{l} \mathsf{X}_{l}^{\top}$, can be efficiently precomputed and cached for reuse in subsequent optimization steps, improving computational efficiency in practice.

Existing PTQ methods typically use one of two possible forms for activation inputs $\mathsf{X}_{l}$:
Either quantized activations $\B{X}_{l}$, obtained by forward propagating the calibration dataset through previously quantized weights $\{\widehat{\B{W}}_{1}, \ldots, \widehat{\B{W}}_{l-1}\}$, or full-precision activations $\B{X}_{l}$, resulting from forward propagation through the original, unquantized weights $\{\B{W}_{1}, \ldots, \B{W}_{l-1}\}$.
There is no consensus among existing PTQ methods \citep{frantar2022gptq, lin2024awq, chee2023quip} regarding whether quantized or full-precision activations produce better quantization outcomes.

Leading layer-wise PTQ methods use distinct optimization strategies to approximate the behavior of the original model while adhering to the foundational sequential layer-wise framework in Eq.~\eqref{eq:original-layerwise-general}.
GPTQ~\citep{frantar2022gptq}, for example, uses quantized activations, $\mathsf{X}_{l} = \widehat{\B{X}}_{l}$, and quantizes parameters row-wise by sequentially minimizing reconstruction error and correcting residuals in the remaining unquantized entries until each row is fully quantized. 
AWQ~\citep{lin2024awq} uses original activations, $\mathsf{X}_{l} = \B{X}_{l}$, and identifies a small subset of \emph{salient weights} whose magnitudes significantly influence the layer outputs, subsequently rescaling these weights before quantization.

\section{Bottleneck: Quantization Error Accumulation and Growth}\label{sec:bottleneck}
\begin{figure}
    \centering
    \includegraphics[width=\linewidth]{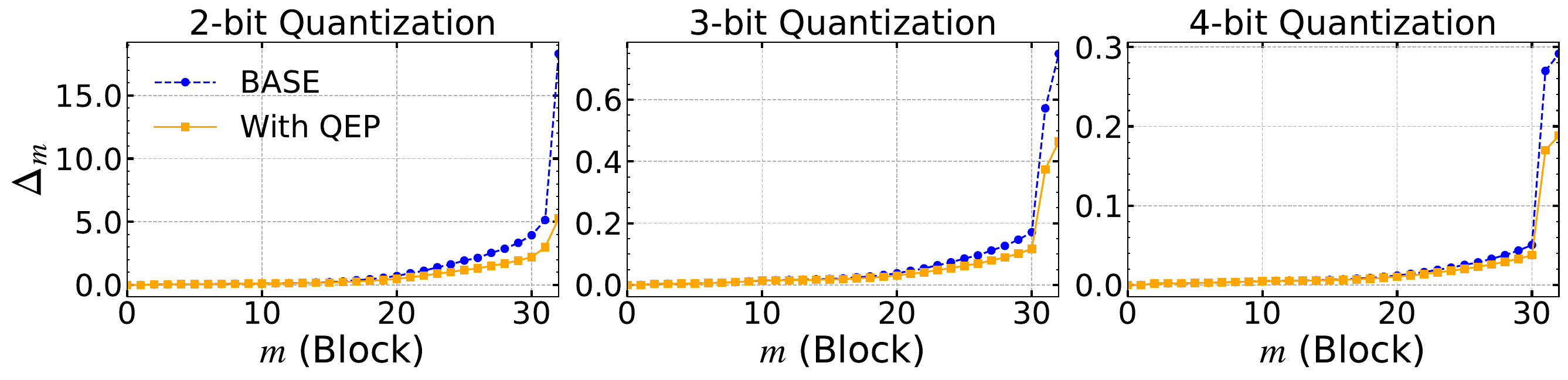}
    \caption{
    Accumulation and growth of quantization errors across layers in a partially quantized Llama2-7B model \citep{touvron2023llama}.
    The first $10$ Transformer blocks are quantized using standard RTN (BASE) and QEP-enhanced RTN (With QEP), while the remaining Transformer blocks after the $10$th remain at full precision.
    The plot shows the squared Frobenius norm $\Delta_{m}$, defined in Eq.~\eqref{eq:metric-of-qe-growth}, between the original and partially quantized outputs at each Transformer block $m$.
    }
    \label{fig:bottleneck}
\end{figure}

To motivate our proposed approach, this section first revisits the core layer-wise optimization formulation given by Eq.~\eqref{eq:original-layerwise-general}, emphasizing its key limitation: 
The \emph{accumulation} and \emph{growth} of quantization errors across layers significantly degrade the performance.
We investigate this phenomenon using experiments conducted on the pre-trained Llama-2-7B model~\citep{touvron2023llama}.
Specifically, we quantize only the first $10$ Transformer blocks~\citep{vaswani2017attention}, while keeping all subsequent blocks in full precision.
To quantify the propagation and accumulation of the errors, we measure the discrepancy between fully precise and partially quantized outputs at each block using a calibration dataset.
Let $\mathtt{TransBlock}_{m}(\cdot)$ denote the original full-precision $m$-th Transformer block, and $\widehat{\mathtt{TransBlock}}_{m}(\cdot)$ denote its quantized counterpart.
We evaluate the following metric at the $m$-th block:
\begin{equation}
    \label{eq:metric-of-qe-growth}
    \Delta_{m} = \left\| f_{m}(\B{X}) - \widehat{f}_m(\B{X}) \right\|_{F}^{2},
\end{equation}
\begin{align}
    &f_{m}(\B{X}) \coloneqq \mathtt{TransBlock}_{m} \circ \cdots \circ \mathtt{TransBlock}_{n+1} \circ \mathtt{TransBlock}_{n} \circ \cdots \circ \mathtt{TransBlock}_{1}(\B{X}), \\
    &\widehat{f}_{m}(\B{X}) \coloneqq \mathtt{TransBlock}_{m} \circ \cdots \circ \mathtt{TransBlock}_{n+1} \circ \widehat{\mathtt{TransBlock}}_{n} \circ \cdots \circ \widehat{\mathtt{TransBlock}}_{1}(\B{X}).
\end{align}
This experiment sets $n=10$.
Figure~\ref{fig:bottleneck} shows an approximately exponential \emph{accumulation} or errors within the quantized layer, as well as an error \emph{growth} that persists in the unquantized layers. 
This \emph{growth} occurs due to the layer-wise \emph{independent} quantization approach described in Eq.~\eqref{eq:original-layerwise-general}, which neither accounts for quantization error propagated from previous layers nor corrects previously accumulated errors, thus exacerbating error growth in subsequent unquantized layers. 
The exponential accumulation of quantization errors observed empirically can also be theoretically explained under mild conditions, as detailed in Appendix~\ref{app-subsec:error-exponential}.
Therefore, instead of treating layer-wise quantization as a series of independent optimization problems, it is essential to reformulate the original layer-wise optimization presented in Eq.~\eqref{eq:original-layerwise-general} to mitigate error accumulation and growth.

\section{QEP: Quantization Error Propagation}\label{sec:qep}
Existing layer-wise independent PTQ has inherent limitations, particularly the \emph{accumulation} and \emph{growth} of quantization errors discussed in Section \ref{sec:bottleneck}. 
To address these limitations, we introduce \underline{\textbf{Q}}uantization \underline{\textbf{E}}rror \underline{\textbf{P}}ropagation (\textbf{QEP}), a general, lightweight, and scalable framework that improves layer-wise PTQ by propagating quantization errors.
In subsequent sections, we provide theoretical evidence showing that QEP effectively reduces quantization errors.

\subsection{Problem Reformulation}\label{subsec:problem-reformulation}
We reformulate the layer-wise \emph{independent} optimization strategy presented in Eq.~\eqref{eq:original-layerwise-general} to propagate quantization errors across layers effectively.
Instead of minimizing output differences based on shared input activations $\mathsf{X}_l$, our reformulation directly minimizes the discrepancy between full-precision and quantized outputs, each computed using their respective upstream inputs. 
Formally, for each layer $l$, we optimize the discrete quantized weight matrix $\widehat{\B{W}}_{l}$ as follows:
\begin{equation}
    \label{eq:form-qep}
    \min_{\widehat{\B{W}}_{l} \in \mab{Q}^{n_l \times d_l}}
    \left\| \B{W}_l \B{X}_l - \widehat{\B{W}}_l \widehat{\B{X}}_l \right\|_{F}^{2}.
\end{equation}
This objective ensures that the quantized weights $\widehat{\B{W}}_{l}$ are optimized not only to independently approximate the full-precision weights $\B{W}_{l}$ but also to counteract and compensate for the cumulative quantization errors introduced by previous layers.
In contrast to the existing objective in Eq.~\eqref{eq:original-layerwise-general}, where the trivial optimal solution is $\widehat{\B{W}}_{l} = \B{W}_{l}$ if $\B{W}_{l} \in \mab{Q}^{n_{l} \times d_{l}}$, the optimal solution under the formulation in Eq.~\eqref{eq:form-qep} is generally $\widehat{\B{W}}_{l} \neq \B{W}_{l}$, explicitly enabling error correction and accounting for accumulated quantization errors.

Although the modification from Eq.~\eqref{eq:original-layerwise-general} seems straightforward, Eq.~\eqref{eq:form-qep} inherently breaks the key structural simplification that facilitates efficient quantization in existing PTQ frameworks.
Specifically, the optimization in Eq.~\eqref{eq:form-qep} no longer solely depends on the Hessian matrix $\mathsf{H}_{l}$, thereby preventing the direct use of existing Hessian-based acceleration methods for quantization.
In the following section, we address this challenge by proposing a practical and efficient weight correction scheme that overcomes this limitation while retaining the advantages of our error-propagation approach.

\subsection{Weight Correction}\label{subsec:weight-correction}
To efficiently perform quantization by the objective in Eq.~\eqref{eq:form-qep} as in existing layer-wise PTQ methods, we relax the discrete feasible set to a continuous domain, leading to the following proposition.
\begin{proposition}
    \label{prop:optimal-w}
    Assume that the matrix $\widehat{\B{H}}_l$ is invertible. Then, after relaxing the discrete feasible set $\mab{Q}^{n_{l} \times d_{l}}$ into the continuous domain $\mab{R}^{n_{l} \times d_{l}}$, the optimal solution $\B{W}_{l}^{\ast}$ is given by the following closed-form expression:
    \begin{equation}
        \label{eq:correction-weight}
         \B{W}_{l}^{\ast}\coloneqq \B{W}_{l} + \B{W}_{l} \B{\delta}_{l} \widehat{\B{X}}_{l}^{\top} \widehat{\B{H}}_{l}^{-1}
        = \argmin_{\widehat{\B{W}}_l \in \mab{R}^{n_l \times d_l}} 
        \left\| \B{W}_l \B{X}_l - \widehat{\B{W}}_l \widehat{\B{X}}_l \right\|_F^2,
    \end{equation}
    where $\B{\delta}_l \coloneqq \B{X}_{l} - \widehat{\B{X}}_{l}$ represents the accumulated quantization error from proceeding layers, $\widehat{\B{H}}_l \coloneqq \widehat{\B{X}}_l \widehat{\B{X}}_l^\top$ denotes the empirical Hessian constructed from quantized activations.
\end{proposition}
The proof of Proposition~\ref{prop:optimal-w} is provided in Appendix~\ref{sec:derivation}.
Proposition~\ref{prop:optimal-w} highlights an important distinction from the existing formulation given by Eq.~\eqref{eq:original-layerwise-general}.
Specifically, when upstream quantization introduces non-negligible errors, i.e., $\B{\delta}_{l-1} \neq 0$, the optimal quantized weights differ from straightforward approximations of the original weights $\B{W}_{l}$.
Instead, the optimal solution explicitly includes a correction term that compensating for accumulated quantization errors.

This corrected weight enables us to reformulate the equivalent optimization objective within the original discrete set $\widehat{\B{W}}_{l} \in \mab{Q}^{n_{l} \times d_{l}}$ as follows:
\begin{equation}
    \label{eq:correction-optimization-problem}
    \min_{\widehat{\B{W}}_l \in \mab{Q}^{n_l \times d_l}} 
    \left\| \B{W}_l^\ast \widehat{\B{X}}_l - \widehat{\B{W}}_l \widehat{\B{X}}_l \right\|_F^2.
\end{equation}
This objective shares the same structure as Eq.~(\ref{eq:original-layerwise-general}), with $\B{W}_{l}$ replaced by its corrected counterpart $\B{W}_{l}^{\ast}$.
This reformulation restores the quadratic structure found in Eq.~\eqref{eq:original-layerwise-general}, facilitating efficient optimization through the Hessian matrix $\mathsf{H}_{l} = \widehat{\B{H}}_l$.
The structure of Eq.~\eqref{eq:correction-optimization-problem} allows for seamless integration with various existing layer-wise PTQ methods, as discussed in Section\ref{sec:related-work}.
Furthermore, the proposed layer-wise quantization formulation in Eq.~\eqref{eq:form-qep} formally guarantees improved quantization accuracy compared to the existing layer-wise \emph{independent} PTQ defined in Eq.~\eqref{eq:original-layerwise-general}. 
Specifically, we establish the following theoretical result:
\begin{theorem}[Informal]
\label{theorem:gurantee-reduction}
Consider an $L$-layer neural network defined by:
\begin{equation}
    f_{\B{\theta}}(X) = \sigma_{L}(\B{W}_L  \sigma_{L-1}(\B{W}_{L-1}\cdots \sigma_{2}(\B{W}_2  \sigma_{1}(\B{W}_1 \B{X}))\cdots)),
\end{equation}
where each activation function $\sigma_{l}$ is Lipschitz continuous and $\B{\theta}$ denotes the set of all full-precision parameters $\{\B{W}_l\}_{l=1}^{L}$.
The output quantization error of the proposed quantization method defined in Eq.~\eqref{eq:form-qep} is bounded by that of the existing layer-wise PTQ defined in Eq.~\eqref{eq:original-layerwise-general}:
\begin{equation}
    \left\|f_{\theta}(\B{X})-f_{\widehat{\B{\theta}}_{\mathrm{QEP}}}(\B{X})\right\|_F 
    \le 
    \left\|f_{\B{\theta}}(\B{X})-f_{\widehat{\B{\theta}}_{\mathrm{BASE}}}(\B{X}) \right\|_F.
\end{equation}
where $\widehat{\B{\theta}}_{\mathrm{QEP}}$ and $\widehat{\B{\theta}}_{\mathrm{BASE}}$ denote the sets of parameters quantized by the objective in Eq.~\eqref{eq:form-qep} and the base PTQ method by the objective in Eq.~\eqref{eq:original-layerwise-general}, respectively.
\end{theorem}
Explicit conditions and detailed proof are provided in Appendix~\ref{subsec:gurantee-reduce}. 
The additional computational overhead arises solely from computing the correction term $\B{\delta}_{l} \widehat{\B{X}}_{l}^{\top}$, since computing the Hessian inverse $\widehat{\B{H}}_{l}^{-1}$ remains unchanged from existing layer-wise \emph{independent} PTQ.
As empirically demonstrated in Section~\ref{subsec:results-llama2}, this additional computation requires significantly less runtime compared to the quantization processes of layer-wise PTQ methods, even for large-scale LLMs, due to the tunable mechanism described in the next section.

\subsection{Controlling Propagation Strength}\label{subsec:controlling-propagation}
Although solving Eq.~\eqref{eq:correction-optimization-problem} effectively reduces the accumulation of quantization error, it can lead to overfitting.
This issue is particularly pronounced when the calibration dataset is small and insufficiently representative of the target task, or when the model includes blocks with a large number of parameters such as the MLP blocks commonly found in transformer architectures, causing the correction to overfit the calibration dataset.

To address this issue, we introduce a tunable propagation mechanism that generalizes the correction term using a scaling parameter $\alpha_{l} \in [0, 1]$:
\begin{equation}
    \label{eq:alpha-corrected-weight}
    \B{W}_{l}^{\ast}(\alpha_{l}) = \B{W}_{l} + \alpha_{l} \B{W}_{l} \B{\delta}_{l} \widehat{\B{X}}_{l}^{\top} \widehat{\B{H}}_{l}^{-1}.
\end{equation}
Here, setting $\alpha_{l} = 1$ recovers original fully-corrected case presented in Eq.~\eqref{eq:correction-weight}, whereas setting $\alpha_{l} = 0$ corresponds to the existing approach in Eq.~\eqref{eq:original-layerwise-general} under the setting that $\mathsf{X}_{l} = \widehat{\B{X}}_{l}$.
This tunable correction mechanism relates to the following regularization optimization:
\begin{proposition}\label{prop:alpha_lambda_correspondence}
The parameter $\alpha_{l}$ corresponds to the regularization parameter $\lambda$ in the following optimization problem:
\begin{equation}\label{eq:ridge_obj}
    \min_{\widehat{\B{W}}_{l} \in \mab{Q}^{n_{l} \times d_{l}}}\|\B{W}_{l}\B{X}_{l}-\widehat{\B{W}}_{l}\widehat{\B{X}}_{l}\|_{F}^{2}
    +\lambda_{l}\|\B{W}_{l}-\widehat{\B{W}}_{l}\|_{F}^{2},~~ \lambda_{l} \in \mab{R}_{+}.
\end{equation}
Specifically, as $\alpha_{l}$ increases from $0$ to $1$, the corresponding parameter $\lambda_{l}$ decreases from $+\infty$ to $0$.
\end{proposition}
The derivation is provided in Appendix~\ref{sec:alpha_lambda_equiv}.
Additionally, the following proposition is established.
\begin{proposition}
    \label{cor:alpha-monotonicity}
    Under the same assumptions in Theorem \ref{theorem:gurantee-reduction}, the output quantization error of the method employing QEP with parameter $\{\alpha_{l}\}_{l=1}^{L}$ decreases monotonically as each $\alpha_l$ approaches $1$.
\end{proposition}
Explicit conditions and comprehensive proofs of this proposition are provided in Appendix~\ref{subsec:gurantee-reduce}.
Consequently, the parameter $\alpha_{l}$ effectively controls overfitting, analogous to regularization techniques, and importantly provides a systematic way to balance overfitting and underfitting in layer-wise PTQ methods.
Indeed, this parameter is crucial for preventing overfitting, especially in MLP blocks, which contain more parameters than other blocks.

Furthermore, in large-scale LLMs, the high-dimensional activations in MLP layers often result in computationally expensive correction terms.
In these cases, selectively setting $\alpha_{l} = 0$ for specific layers eliminates the computational cost of the correction term and acts as implicit regularization, potentially improving generalization.
Therefore, appropriately setting $\alpha_{l} = 0$ can reduce the correction time by approximately one-third and one-half.
Developing adaptive strategies for layer-wise, data-aware, or resource-efficient tuning of $\alpha_{l}$ is a promising direction for future research.
In the following, we refer to the overall approach, including the tunable mechanism controlled by $\{\alpha_{l}\}_{l=1}^{L}$, as \underline{\textbf{Q}}uantization \underline{\textbf{E}}rror \underline{\textbf{P}}ropagation (\textbf{QEP}).

\section{Experiments}\label{sec:experiments}

We conduct experiments to validate the effectiveness of QEP in improving the performance of layer-wise PTQ relative to existing methods.

\paragraph{Baselines}
We use representative layer-wise PTQ methods based on linear quantization such as round-to-nearest (RTN)~\citep{frantar2022gptq, dettmers2023case}, GPTQ~\citep{frantar2022gptq}, AWQ~\citep{lin2024awq}, and QuIP~\citep{chee2023quip}.
Although previous studies have explored extensions, such as non-linear and block-wise quantization, as discussed in Section \ref{sec:related-work}, these techniques are orthogonal to the core improvement introduced by QEP. Therefore, to isolate and emphasize the impact of QEP, we focus on these representative layer-wise PTQ methods.

\paragraph{Quantization}
This study focuses on weight-only quantization schemes, specifically per-channel and group-wise quantization, which have recently shown superior trade-offs between efficiency and accuracy~\citep{dettmers2023case, frantar2022gptq, lin2024awq}. 
The main text evaluates per-channel quantization under INT4, INT3, and INT2 precision settings. 
Due to space constraints, detailed results for group-wise quantization are presented in Appendix \ref{sec:add-experiments}.
For the propagation strength parameter $\alpha_{l}$, we adopt a representative default value of $\alpha_{l}=\nicefrac{1}{2}$ for all layers, except for the MLP layers in the Llama-2 70B model, for which we set $\alpha_{l}=0$.
Tuning $\alpha_{l}$ can further improve performance but is beyond the scope of this study and is left for future work.

\paragraph{Datasets}
Following previous studies, we evaluate the Hessian matrix using the same default calibration datasets used in their original implementations.
Specifically, GPTQ and QuIP use the C4 dataset~\citep{frantar2022gptq} for calibration, while AWQ uses the Pile dataset~\citep{gao2020pile}.
Following \citet{frantar2022gptq}, we evaluate the correction term in Eq.~\eqref{eq:correction-weight} using $128$ randomly sampled segments of $2048$ tokens each from the C4 dataset\citep{raffel2020exploring}, which consists of web-crawled text excerpts.

\paragraph{Models}
Following \citet{lin2024awq, frantar2022gptq}, we evaluate our method on recent popular LLMs, namely the Llama-2 and Llama-3 model families~\citep{touvron2023llama}, with size ranging from 7B to 70B parameters, as well as Mistral-7 B~\citep {jiang2024identifying}.
These models demonstrate superior performance compared to other open-source LLMs~\citep{zhang2022opt, scao2022bloom} and have become widely adopted as foundational models for numerous derivative open-source models~\citep{taori2023stanfordalpaca, vicuna2023}.

\paragraph{Evaluations}

Following established evaluation protocols from prior studies~\citep{dettmers2022gpt3, xiao2023smoothquant, frantar2022gptq, dettmers2023case, yao2022zeroquant}, we evaluate the quantized LLMs using the perplexity (PPL) on WikiText2~\citep{merity2016pointer}, Penn Treebank (PTB) \citep{marcus-etal-1994-penn}, and C4 \citep{raffel2020exploring}, and zero-shot accuracy on benchmarks including ARC Easy (ArcE) \citep{Boratko2018}, PiQA \citep{bisk2020piqa}, and StoryCloze (SC) \citep{mostafazadeh-etal-2016-corpus}.
Due to space limitations, detailed results for each dataset are provided in Appendix~\ref{sec:add-experiments}.
All experiments are conducted using a single NVIDIA V100 GPU.

\subsection{Results}\label{subsec:results-llama2}

\paragraph{Perplexity}
\begin{table}[tb]
  \centering
  \scriptsize
  \footnotesize
  \caption{Evaluation of perplexities ($\downarrow$) for Llama models on WikiText-2 under various layer-wise PTQ methods and bitwidths.}
  \label{tab:gqep_perplexity}
  \setlength{\tabcolsep}{4pt}
  \renewcommand{\arraystretch}{1.12}

  \begin{tabular}{c|cc|ccccc}
    \toprule
    \textbf{Bits} & \textbf{Method} & \textbf{QEP} &
    \textbf{Llama-2-7B} & \textbf{Llama-2-13B} & \textbf{Llama-2-70B} 
    & \textbf{Llama-3-8B} & \textbf{Mistral-7B}\\
    \midrule
    FP16 & - & - & 5.472 & 4.883 & 3.319 & 6.137 & 5.255
    \\
    \midrule
    \multirow{8}{*}{INT4}
      & \multirow{2}{*}{RTN}  & \xmark & 6.116 & 5.206 & 3.672 & 8.540 & 5.997 \\
      &                       & \cmark & \textbf{\q{6.017}} & \textbf{\q{5.165}} & \textbf{\q{3.621}} & \textbf{\q{8.021}} & \textbf{\q{5.877}} \\ 
      & \multirow{2}{*}{GPTQ} & \xmark & 6.083 & 5.167 & 3.594 & 147.912 & 5.643 \\
      &                       & \cmark & \textbf{\q{5.933}} & \textbf{\q{5.127}} & \textbf{\q{3.576}} & \textbf{\q{9.509}} & \textbf{\q{5.528}} \\ 
      & \multirow{2}{*}{AWQ}  & \xmark & 5.831 & 5.064 & 3.484 & 7.108 & 5.716 \\
      &                       & \cmark & \textbf{\q{5.756}} & \textbf{\q{5.041}} & \textbf{\q{3.479}} & \textbf{\q{6.981}} & \textbf{\q{5.636}} \\ 
      & \multirow{2}{*}{QuIP} & \xmark & 8.434 & 5.137 & 3.826 & 6.998 & 11.109 \\
      &                       & \cmark & \textbf{\q{5.753}} & \textbf{\q{5.034}} & \textbf{\q{3.485}} & \textbf{\q{6.650}} & \textbf{\q{5.479}} \\
    \midrule
    \multirow{8}{*}{INT3}
      & \multirow{2}{*}{RTN}  & \xmark & 539.866 & 10.688 & 7.530 & 2276.227 & 29.390 \\
      &                       & \cmark & \textbf{\q{17.309}} & \textbf{\q{7.458}} & \textbf{\q{5.648}} & \textbf{\q{86.430}} & \textbf{\q{10.241}} \\
      & \multirow{2}{*}{GPTQ} & \xmark & 10.881 & 6.632 & 4.860 & 64.457 & 8.247 \\
      &                       & \cmark & \textbf{\q{7.898}} & \textbf{\q{6.245}} & \textbf{\q{4.102}} & \textbf{\q{18.845}} & \textbf{\q{7.347}} \\ 
      & \multirow{2}{*}{AWQ}  & \xmark & 15.299 & 6.448 & 4.362 & 11.802 & 7.902 \\
      &                       & \cmark & \textbf{\q{11.131}} & \textbf{\q{6.092}} & \textbf{\q{4.103}} & \textbf{\q{10.713}} & \textbf{\q{7.169}} \\ 
      & \multirow{2}{*}{QuIP} & \xmark & 12.048 & 5.503 & 4.135 & 8.288 & 7.108\\
      &                       & \cmark & \textbf{\q{6.154}} & \textbf{\q{5.347}} & \textbf{\q{3.813}} & \textbf{\q{7.703}} & \textbf{\q{5.842}} \\
    \midrule
    \multirow{8}{*}{INT2}
      & \multirow{2}{*}{RTN}  & \xmark & \textbf{17783.918} & \textbf{51152.832} & 26077.172 & 1437176.750 & 78488.328\\
      &                       & \cmark & \q{97153.266} & \q{61158.555} & \textbf{\q{26063.672}} & \textbf{\q{554142.313}} & \textbf{\q{50540.059}}\\ 
      & \multirow{2}{*}{GPTQ} & \xmark & 13051.469 & \textbf{1301.395} & 107.458 & \textbf{236596.891} & 3543.708 \\
      &                       & \cmark & \textbf{\q{7214.328}} & \q{2782.353} & \textbf{\q{52.472}} & \q{282245.188} & \textbf{\q{1665.287}} \\ 
      & \multirow{2}{*}{AWQ}  & \xmark & \textbf{199448.797} & 93036.517 & \textbf{81834.344} & 1044956.250 & \textbf{31391.543} \\
      &                       & \cmark & \q{229888.406} & \textbf{\q{74735.836}} & \q{88684.156} & \textbf{\q{639158.313}} & \q{32668.666}\\ 
      & \multirow{2}{*}{QuIP} & \xmark & 65.593 & 11.232 & 6.536 & 70.518 & 26.632 \\
      &                       & \cmark & \textbf{\q{11.972}} & \textbf{\q{8.417}} & \textbf{\q{5.869}} & \textbf{\q{27.326}} & \textbf{\q{9.586}} \\
    \bottomrule
  \end{tabular}
\end{table}
Table \ref{tab:gqep_perplexity} summarizes PPL results of various quantized models evaluated on WikiText2, comparing several bit-widths and different layer-wise PTQ methods, both with and without QEP. 
Additional C4 and PTB dataset results are provided in Appendix \ref{app-subsec:add-perplexity-results}, demonstrating consistent trends in the following.
Our results indicate that incorporating QEP significantly enhances the performance of layer-wise PTQ, substantially reducing perplexity across nearly all tested methods and quantization levels. 
In medium-bit scenarios such as INT4 and INT3, where AWQ already exhibits strong performance, applying QEP yields further improvements. 
At 2-bit quantization, existing layer-wise PTQ methods based on linear quantization typically suffer severe PPL degradation, rendering deployment infeasible. 
However, QEP effectively mitigates this issue, making INT2 quantization achievable with practical perplexity levels. 
Notably, QEP-enhanced QuIP achieves state-of-the-art perplexity results among all tested layer-wise PTQ methods. 
Similar significant improvements are observed for RTN, GPTQ, and AWQ at INT2g32, INT2g64, and INT2g128 quantization levels; see Appendix \ref{app-subsec:add-perplexity-results} for details.

\paragraph{Zero-shot tasks}

\begin{table}[tb]
  \centering
  \footnotesize
  \caption{Zero-shot average accuracy ($\uparrow$) on ARC-Easy, PIQA, and StoryCloze for Llama models across three quantization settings.}
  \setlength{\tabcolsep}{4pt}
  \renewcommand{\arraystretch}{1.12}
  \label{tab:qep_average}
  \begin{tabular}{c|cc|ccccc}
    \toprule
    \textbf{Bits} & \textbf{Method} & \textbf{QEP} &
    \textbf{Llama-2-7B} & \textbf{Llama-2-13B} & \textbf{Llama-2-70B} 
    & \textbf{Llama-3-8B} & \textbf{Mistral-7B}  \\
    \midrule
    FP16 & - & - & 0.7601 & 0.7840 & 0.8014 & 0.7920 & 0.8056
    \\
    \midrule
    \multirow{8}{*}{INT4}
      & \multirow{2}{*}{RTN}  & \xmark & 0.6802 & \textbf{0.7160} & 0.7325 & 0.7643 & 0.7831 \\
      &                       & \cmark & \textbf{\q{0.6844}} & \q{0.7131} & \textbf{\q{0.7343}} & \textbf{\q{0.7686}} & \textbf{\q{0.7921}} \\
      & \multirow{2}{*}{GPTQ} & \xmark & \textbf{0.6817} & \textbf{0.7134} & 0.7306 & 0.4812 & \textbf{0.7906} \\
      &                       & \cmark & \q{0.6795} & \q{0.7104} & \textbf{\q{0.7308}} & \textbf{\q{0.7531}} & \q{0.7904}\\
      & \multirow{2}{*}{AWQ}  & \xmark & 0.6832 & 0.7120 & 0.7257 & 0.7821 & 0.7956 \\
      &                       & \cmark & \textbf{\q{0.6870}} & \textbf{\q{0.7126}} & \textbf{\q{0.7331}} & \textbf{\q{0.7879}} & \textbf{\q{0.7967}} \\
      & \multirow{2}{*}{QuIP} & \xmark & 0.6500 & \textbf{0.7248} & 0.7285 & \textbf{0.7872} & 0.7204 \\
      &                       & \cmark & \textbf{\q{0.6920}} & \q{0.7167} & \textbf{\q{0.7311}} & \q{0.7800} & \textbf{\q{0.8012}} \\
    \midrule
    \multirow{8}{*}{INT3}
      & \multirow{2}{*}{RTN}  & \xmark & 0.4770 & 0.6082 & 0.6402 & 0.4560 & 0.6448 \\
      &                       & \cmark & \textbf{\q{0.5802}} & \textbf{\q{0.6550}} & \textbf{\q{0.6939}}  & \textbf{\q{0.5388}} & \textbf{\q{0.6963}} \\
      & \multirow{2}{*}{GPTQ} & \xmark & 0.6367 & 0.6747 & 0.7043 & 0.4891 & 0.7305 \\
      &                       & \cmark & \textbf{\q{0.6549}} & \textbf{\q{0.6853}} & \textbf{\q{0.7078}} & \textbf{\q{0.5901}} & \textbf{\q{0.7422}}\\
      & \multirow{2}{*}{AWQ}  & \xmark & 0.5840 & 0.6886 & 0.7209 & 0.7074 & 0.7534 \\
      &                       & \cmark & \textbf{\q{0.6264}} & \textbf{\q{0.6916}} & \textbf{\q{0.7283}} & \textbf{\q{0.7216}} & \textbf{\q{0.7675}} \\
      & \multirow{2}{*}{QuIP} & \xmark & 0.6232 & 0.7034 & 0.7246 & 0.7433 & 0.7422 \\
      &                       & \cmark & \textbf{\q{0.6804}} & \textbf{\q{0.7128}} & \textbf{\q{0.7273}} & \textbf{\q{0.7549}} & \textbf{\q{0.7933}} \\
    \midrule
    \multirow{8}{*}{INT2}
      & \multirow{2}{*}{RTN}  & \xmark & 0.4139 & \textbf{0.4283} & 0.4147 & \textbf{0.4183} & \textbf{0.4130} \\
      &                       & \cmark & \textbf{\q{0.4199}} & \q{0.4191} & \textbf{\q{0.4145}} & \q{0.4108} & \q{0.4084}\\
      & \multirow{2}{*}{GPTQ} & \xmark & 0.4162 & 0.4222 & 0.4356 & 0.4116 & \textbf{0.4159} \\
      &                       & \cmark & \textbf{\q{0.4263}} & \textbf{\q{0.4283}} & \textbf{\q{0.4714}} & \textbf{\q{0.4228}} & \q{0.4148} \\
      & \multirow{2}{*}{AWQ}  & \xmark & \textbf{0.4213} & \textbf{0.4176} & 0.4129 & \textbf{0.4164} & 0.4177 \\
      &                       & \cmark & \q{0.4162} & \q{0.4165} & \textbf{\q{0.4140}} & \q{0.4150} & \textbf{\q{0.4181}} \\
      & \multirow{2}{*}{QuIP} & \xmark & 0.4667 & 0.5945 & 0.6628 & 0.4600 & 0.5422 \\
      &                       & \cmark & \textbf{\q{0.5926}} & \textbf{\q{0.6404}} & \textbf{\q{0.6998}} & \textbf{\q{0.5121}} & \textbf{\q{0.6858}}\\
    \bottomrule
  \end{tabular}
\end{table}

We evaluate the zero-shot accuracy of quantized models on several tasks.
Table \ref{tab:qep_average} summarizes the average accuracy for the ArcE, PiQA, and SC datasets. 
Detailed results for each dataset are provided in Appendix \ref{app-subsec:detailed-acc-individual-task}.
Consistent with the perplexity results, QEP effectively improves existing layer-wise PTQ methods.
Notably, the performance gains from QEP are especially pronounced with INT2 quantization. 
For Llama-2-70B, the QEP-enhanced QuIP at INT2 achieves performance comparable to RTN and GPTQ at INT3 quantization.

\begin{wraptable}{r}{0.4\textwidth}
    \vspace{-20pt}
    \centering
    \small
    \caption{Runtime comparison of the quantization process.}
    \label{tab:runtime}
    \begin{tabular}{l|ccc}
        \toprule
        Runtime & \multicolumn{3}{c}{Llama-2} \\
                & 7B & 13B & 70B \\
        \midrule
        GPTQ & 14.9m & 26.4m & 2.9h\\
        AWQ & 13.6m & 25.4m & 2.4h\\
        
        QEP + RTN & 10.9m & 19.6m & 1.7h \\
        \bottomrule
    \end{tabular}
    \vspace{-10pt}
\end{wraptable}

\paragraph{Runtime}
We examine the impact of computation time required for the correction term.
Table \ref{tab:runtime} shows the processing time of each layer-wise PTQ.
Since the quantization processing time for RTN is only a few seconds and thus negligible, the measured time for QEP+RTN is primarily due to computing the preprocessing of the correction term.
This result indicates that calculating the QEP correction term requires significantly less computation time than other existing layer-wise PTQ quantization processes.
Moreover, using the same calibration dataset for weight correction and quantization reduces preprocessing overhead by approximately one-half to one-third by reusing computational steps.

\paragraph{Robustness}
\begin{wraptable}{r}{0.46\textwidth} 
    \vspace{-28pt}
    \centering
    \small
    \caption{Perplexity relative to RTN on WikiText2, comparing GPTQ and QEP+RTN calibrated with C4, PTB, and WikiText2.}
    \label{tab:calibaration-sets}
    \begin{tabular}{l|ccc}
        \toprule
        \multirow{2}{*}{PPL to RTN ($\downarrow$)} & \multicolumn{3}{c}{Calibration Dataset} \\
         & C4 & PTB & WikiText2 \\
        \midrule
        GPTQ & -0.25 & +0.07 & -0.46 \\
        QEP + RTN & \q{\textbf{-0.33}} & \q{\textbf{-0.30}} & \q{\textbf{-0.49}} \\
        \bottomrule
    \end{tabular}
    \vspace{-15pt}
\end{wraptable}
As discussed in Section \ref{subsec:controlling-propagation}, our method adaptively controls propagation strength in Eq.~\eqref{eq:alpha-corrected-weight} to mitigate overfitting to the calibration dataset. 
In this section, we empirically validate this approach.
Table \ref{tab:calibaration-sets} compares the perplexity difference among QEP-enhanced RTN, GPTQ, and RTN when quantizing Llama-2-7B, evaluated on WikiText2 across various calibration datasets.
Consistent with prior findings \citep{lin2024awq}, GPTQ exhibits significant sensitivity to the calibration dataset: it outperforms RTN on C4 and WikiText2 but experiences notable performance degradation on PTB.
In contrast, QEP-enhanced RTN consistently improves performance across all calibration datasets, demonstrating robustness to distributional shifts. 
This highlights the effectiveness of propagation control in preventing overfitting to the calibration dataset.

\vspace{-5pt}
\section{Conclusion}\label{sec:conclusion}

We revisit the core design of layer-wise PTQ and identify a critical limitation: the exponential accumulation and growth of quantization errors across network layers.
To address this issue, we propose QEP, a general framework that explicitly propagates and compensates for accumulated quantization errors.
Extensive experiments demonstrate that QEP substantially improves performance, especially in low-bit quantization scenarios.
These findings underscore that meaningful progress in layer-wise PTQ can still be made by revisiting fundamental strategies, complementing recent trends primarily centered around non-linear and block-wise quantization techniques.
Integrating QEP with these advanced quantization methods in the future presents a promising approach toward achieving extreme compression, potentially exceeding QAT performance.

\paragraph{Limitations}
QEP relies on a small calibration set, as in other layer-wise PTQ approaches, which makes performance sensitive to data quality; however, it overfits less than comparable methods such as GPTQ and AWQ.
The method also introduces a per-layer propagation-strength parameter $\alpha_l$; Although this parameter is tunable, a fixed value, e.g., $\alpha_l=\nicefrac{1}{2}$, works well in most cases, and automatic learning of $\alpha_{l}$ is left for future work.

\newpage       
\bibliographystyle{unsrtnat}
\bibliography{ref}

\newpage
\appendix

\section{Additional Related Work}\label{sec:additional-related work}

\paragraph{Quantization error mitigation}

Defensive Quantization (DQ) \citep{lin2019defensive} mitigates error accumulation by introducing an orthogonality penalty to the weights, which reduces the correlation-driven amplification of quantization noise It also employs gradient-based quantization to optimize the positioning of quantization levels, further suppressing propagated error and enhancing robustness.
In contrast, the QEP module serves as a plug-and-play component for general layer-wise quantization, explicitly modulating propagation strength under standard linear, gradient-free quantization methods, such as GPTQ, AWQ, and QuIP.

\paragraph{Relation to GPTAQ} 
GPTAQ \citep{li2025gptaq} optimizes the same local objective but is closely related to GPTQ, and therefore does not easily generalize to other layer-wise PTQ methods such as AWQ and QuIP. 
In contrast, QEP adds the correction term in Eq.~\eqref{eq:correction-weight} directly to the pre-trained weights, enabling plug-and-play use with diverse PTQ algorithms and ensuring strong performance in low-bit regimes. 
Moreover, whereas GPTAQ offers no guarantee that its local optimization reduces global quantization error, our analysis provides such a guaranty. 
QEP also introduces a per-layer propagation-strength parameter, $\alpha_{l}$, which mitigates the overfitting often observed with GPTQ and GPTAQ.

\paragraph{Extreme low-bit layer-wise PTQ}

In the ultra-low precision regime, the layer-wise PTQ pipeline described in the main text, such as  GPTQ and AWQ with standard linear quantizers, often becomes inadequate. 
As a result, many methods adopt alternative formalisms that introduce additional degrees of freedom beyond naive rounding to avoid catastrophic quality degradation. 
Representative examples include SVID-based 1-bit parameterizations \citep{xu2024onebit}, token-adaptive mixtures of scaling factors \citep{jo2024binarymos}, and structured sparsity designed for extreme quantization \citep{dong2025stbllm}. 
In parallel, \emph{binary-factor} formats decompose each weight matrix into bit-packed sign factors with lightweight diagonal scaling, allowing inference to be largely driven by efficient 1-bit kernels \citep{boza2025dbf, lee2025littlebit}. 
In these formats, optimization is more challenging: strictly binary variables and highly non-smooth objectives can make straight-through estimators brittle \citep{long2021learning, ichikawa2025high, yin2019understanding}, motivating the use of discrete optimization or robust relaxation-based solvers. 
Recent advances in controlled continuous relaxations and annealing-style objectives have emphasized promising techniques such as QQA and iSCO \citep{ichikawa2025continuous, ichikawa2025optimization, sun2023revisiting, ichikawa2024controlling}. 
A natural direction for future work is to examine how these extreme low-bit approaches interact with our error-propagation perspective. For instance, we should investigate whether combining QEP with binary-factor or sparsity-based methods can further suppress cross-layer error growth while maintaining the strong INT2 performance observed here, yielding larger gains as precision approaches 1 bit.

\section{Additional Theoretical Results}\label{sec:add-theoretical-results}
This section presents proofs of the propositions stated in the main text along with additional theoretical analyzes.
To avoid ambiguity, we will fix the following notation throughout this section:
\begin{center}
\renewcommand{\arraystretch}{1.3}
\begin{tabular}{@{}ll@{}}
\toprule
\textbf{Symbol} & \textbf{Description} \\
\midrule
$\B{X} \in \mab{R}^{d_1 \times m}$ & Calibration dataset (input activations at layer 1)\\
$\B{W}_l \in \mab{R}^{n_l \times d_l}$ & Full-precision weight matrix at layer $l$ \\
$\widehat{\B{W}}_l \in \mab{Q}^{n_l \times d_l}$ & Quantized weight matrix at layer $l$ \\
$\sigma_l(\cdot)$ & Activation function at layer $l$ \\
$\B{X}_{l+1} \coloneqq \sigma_l(\B{W}_l\B{X}_l)$ & Full-precision activations at layer $(l+1)$ \\
$\widehat{\B{X}}_{l+1} \coloneqq \sigma_l(\widehat{\B{W}}_l\widehat{\B{X}}_l)$ & Quantized activations at layer $(l+1)$ \\
$\B{\delta}_l \coloneqq \B{X}_l - \widehat{\B{X}}_l$ & Quantization error matrix at layer $l$ \\
$\B{H}_l \coloneqq \B{X}_l \B{X}_l^\top$ & Empirical Hessian of a full-precision model at layer $l$\\
$\widehat{\B{H}}_l \coloneqq \widehat{\B{X}}_l \widehat{\B{X}}_l^\top$ & Empirical Hessian of a quantized model at layer $l$\\
\bottomrule
\end{tabular}
\end{center}
In the following section, we assume that $\B{H}_{l}$ and $\widehat{\B{W}}_{l}$ are invertible.
This assumption is standard in existing layer-wise PTQ methods, which also use these inverse matrices.
To ensure numerical stability, a diagonal matrix $\rho \B{I}$, $\rho > 0$ is commonly added to the Hessian when its inversion becomes numerically unstable.
The subsequent analysis remains consistent and valid even when applying this stabilization procedure, which simply involves adding $\rho \B{I}$ to the Hessian in the following derivations.

Throughout this section, we examine the first-order linear term in the weight
perturbations $\{\B{E}_l\}$. Concretely, we define each quantity as
$\B{A} = \B{A}^{(0)} + \B{A}^{(1)} + \B{R}$, where $\B{A}^{(1)}$ collects all terms linear in
$\{\B{E}_l\}$ and the remainder satisfies
$\|\B{R}\| = \mac{O}(\max_{k}\|\B{E}_k\|_2^2)$ as $\max_k\|\B{E}_k\|_2\to 0$.
This matches practice: with INT8 rounding $\|\B{E}_l\|_F/\|\B{W}_l\|_F = 10^{-2} \sim 10^{-1}$,
so quadratic terms are one order of magnitude smaller than any
first–order contribution.
Furthermore, The baseline PTQ and the QEP pipeline use the same quantiser
configuration, hence they induce \emph{errors of the same order}.
We therefore write the same symbol $\B{E}_l$ for the error matrix
in either scheme; any difference is at most a few percent and does not
affect first–order bounds.

\subsection{Derivation of Proposition \ref{prop:optimal-w}}\label{sec:derivation}
This section presents detailed proofs of Proposition \ref{prop:optimal-w} stated in the main text.

\begin{proof}
First, we rewrite the residual inside the Frobenius norm by using the following relationship: $\B{W}_{l}\B{X}_{l} = \B{W}_{l}\widehat{\B{X}}_{l} + \B{W}_{l}\B{\delta}_{l}$. 
Thus, the objective can be expressed as follows:
\begin{equation}
\left\|
    \B{W}_l\B{X}_l 
    -
    \widehat{\B{W}}_l\widehat{\B{X}}_l
\right\|_F^2
=
\left\|
    (\B{W}_l - \widehat{\B{W}}_l)\widehat{\B{X}}_l
    +
    \B{W}_l\B{\delta}_l
\right\|_F^2.
\end{equation}
Since the objective is a strictly convex quadratic function of $\widehat{\B{W}}_l$
when $\widehat{\B{H}}_l$ is invertible, the stationary point is the unique minimizer.
To find the minimizer $\widehat{\B{W}}_{l}$, we set the gradient of the expression with respect to $\widehat{\B{W}}_l$, equal to zero.  
Using standard matrix calculus, we find that the calculus for a stationary point is
\begin{equation}
(\B{W}_l - \widehat{\B{W}}_l)\widehat{\B{X}}_l \widehat{\B{X}}_l^\top
+
\B{W}_l\B{\delta}_l\widehat{\B{X}}_l^\top
=
\B{0}.
\end{equation}
By defining $\widehat{\B{H}}_l \coloneqq \widehat{\B{X}}_l \widehat{\B{X}}_l^\top$, the above condition can be rewritten as 
\begin{equation}
(\B{W}_l - \widehat{\B{W}}_l)\widehat{\B{H}}_l
=
-\B{W}_l\B{\delta}_l\widehat{\B{X}}_l^\top.
\end{equation}
Assuming $\widehat{\B{H}}_l$ is invertible, we multiply both sides on the right by $\widehat{\B{H}}_l^{-1}$, obtaining
\begin{equation}
\B{W}_l - \widehat{\B{W}}_l
=
-\B{W}_l\B{\delta}_l\widehat{\B{X}}_l^\top
\widehat{\B{H}}_l^{-1},
\end{equation}
and hence
\begin{equation}
\widehat{\B{W}}_l
=
\B{W}_l 
+
\B{W}_l\B{\delta}_l\widehat{\B{X}}_l^\top\widehat{\B{H}}_l^{-1}.
\end{equation}
This closed-form expression is indeed the unique minimizer of the Frobenius norm objective, thus completing the proof.
\end{proof}

\subsection{Quantization Error Accumulation}\label{app-subsec:error-exponential}

This section demonstrates that, under standard layer-wise PTQ, where each layer is quantized independently without considering downstream effects, the activation difference at the output layer, defined as 
$\B{\delta}_{L} \coloneqq \B{X}_{L} - \widehat{\B{X}}_{L}$, grows exponentially with depth, to first order in the quantization noise, under mild conditions.
\begin{proposition}\label{prop:1st-order-expansion-error}
For each layer $l=1,\dots,L+1$, the activation error can be expressed as follows:
\begin{equation}
  \B{\delta}_{l}
  = -\sum_{k=1}^{l-1}
      \left( \prod_{s=k+1}^{l-1} \B{J}_{s} \B{W}_{s} \right) \B{J}_{k} \B{E}_{k} \B{X}_{k}
    + \mathcal{O}\left(\max_{k \le l-1} \|\B{E}_{k}\|_{F}^{2}\right),
\end{equation}
where the empty product $\prod_{s=l}^{l-1}$ is defined to be the identity matrix, and 
$\B{E}_{k} \coloneqq \widehat{\B{W}}_k - \B{W}_k$ represents the weight quantization error at layer $k$.
\end{proposition}

\begin{proof}
Consider explicitly the activations at layer $l$ in both full-precision and quantized forms:
\begin{equation}
\B{X}_l = \sigma_{l-1}(\B{W}_{l-1}\B{X}_{l-1}), 
~~
\widehat{\B{X}}_l = \sigma_{l-1}(\widehat{\B{W}}_{l-1}\widehat{\B{X}}_{l-1}).    
\end{equation}
By recursively applying this relation back to the first layer, we derive the activation difference $\B{\delta}_{l}$ as
\begin{align}
    \B{\delta}_{l} 
    &= \B{X}_{l} - \widehat{\B{X}}_{l}\\
    &= \sigma_{l-1}(\B{W}_{l-1}\B{X}_{l-1}) - \sigma_{l-1}(\widehat{\B{W}}_{l-1}\widehat{\B{X}}_{l-1})\\
    &= \B{J}_{l-1}(\B{W}_{l-1}\B{X}_{l-1}-\widehat{\B{W}}_{l-1}\widehat{\B{X}}_{l-1})
      +\mathcal{O}\left(\max\{\B{E}_{l-1}^{2}, \B{\delta}_{l-1}^{2}\}\right)\\
    &= \B{J}_{l-1}\left[-\B{E}_{l-1}\B{X}_{l-1}+\B{W}_{l-1}\B{\delta}_{l-1}\right] 
      + \mathcal{O}\left(\max\{\B{E}_{l-1}^{2}, \B{\delta}_{l-1}^{2}\}\right)\\
    &= -\B{J}_{l-1}\B{E}_{l-1}\B{X}_{l-1}
       +\B{J}_{l-1}\B{W}_{l-1}\B{\delta}_{l-1}
       + \mathcal{O}\left(\max\{\B{E}_{l-1}^{2}, \B{\delta}_{l-1}^{2}\}\right).
\end{align}
By explicitly expanding $\B{\delta}_{l-1}$, we obtain
\begin{equation}
    \B{\delta}_{l-1} 
    = -\B{J}_{l-2}\B{E}_{l-2}\B{X}_{l-2}
      +\B{J}_{l-2}\B{W}_{l-2}\B{\delta}_{l-2}
      + \mathcal{O}\left(\max\{\B{E}_{l-2}^{2}, \B{\delta}_{l-2}^{2}\}\right).
\end{equation}
Substituting this expression into the previous equation yields
\begin{align}
    \B{\delta}_{l} 
    &= -\B{J}_{l-1}\B{E}_{l-1}\B{X}_{l-1}\\
    &~~+\B{J}_{l-1}\B{W}_{l-1}\left[-\B{J}_{l-2}\B{E}_{l-2}\B{X}_{l-2}
    +\B{J}_{l-2}\B{W}_{l-2}\B{\delta}_{l-2}\right]
    +\mathcal{O}\left(\max\{\B{E}_{l-1}^{2}, \B{E}_{l-2}^{2}, \B{\delta}_{l-2}^{2}\}\right)\\
    &=-\B{J}_{l-1}\B{E}_{l-1}\B{X}_{l-1}
      -\B{J}_{l-1}\B{W}_{l-1}\B{J}_{l-2}\B{E}_{l-2}\B{X}_{l-2}\\
    &~~+\B{J}_{l-1}\B{W}_{l-1}\B{J}_{l-2}\B{W}_{l-2}\B{\delta}_{l-2}
    +\mathcal{O}\left(\max\{\B{E}_{l-1}^{2},\B{E}_{l-2}^{2},\B{\delta}_{l-2}^{2}\}\right).
\end{align}
By recursively repeating this explicit expansion down to the first layer, we obtain the fully expanded form as follows, noting $\B{\delta}_{1} = \B{0}$: 
\begin{equation}
    \B{\delta}_{l} 
    = -\sum_{k=1}^{l-1}\left(\prod_{s=k+1}^{l-1}\B{J}_{s}\B{W}_{s}\right)\B{J}_{k}\B{E}_{k}\B{X}_{k} 
    +\mathcal{O}\left(\max_{k \le l-1}\|\B{E}_{k}\|^{2}\right),
\end{equation}
where the empty product for $s = l, \dots, l-1$ is defined as the identity matrix.
\end{proof}

\begin{proposition}
\label{prop:uniform-bound-for-error}
Assume each activation $\sigma_l:\mab{R}^{n_l\times m}\to\mab{R}^{n_l\times m}$
is $\gamma_l$-Lipschitz with respect to the Frobenius norm and satisfies $\sigma_l(\B{0})=\B{0}$:
\begin{equation}
    \|\sigma_l(\B{U})-\sigma_l(\B{V})\|_F \le \gamma_l\|\B{U}-\B{V}\|_F, 
    ~~\gamma_l>0.
\end{equation}
Let $\B{X}_1=\widehat{\B{X}}_1=\B{X}$; for $l=1,\dots,L-1$ define
\begin{equation}
    \B{X}_{l+1}=\sigma_l(\B{W}_l\B{X}_l),~~
    \widehat{\B{X}}_{l+1}=\sigma_l(\widehat{\B{W}}_l\widehat{\B{X}}_l),~~
    \widehat{\B{W}}_l=\B{W}_l+\B{E}_l,
\end{equation}
and $\B{\delta}_l\coloneq\B{X}_l-\widehat{\B{X}}_l$.
Assume $\|\B{W}_l\|_2>0$ for all $l=1,\dots,L-1$ and set
\begin{equation}
    G_{L-1}\coloneq\prod_{l=1}^{L-1}\gamma_l\|\B{W}_l\|_2,
    ~~
    r\coloneq\max_{1\le k\le L-1}\frac{\|\B{E}_k\|_2}{\|\B{W}_k\|_2}.
\end{equation}
Then the final activation mismatch satisfies the explicit bound
\begin{equation}
    \label{eq:uniform_bound_exact}
    \|\B{\delta}_L\|_F \le \bigl((1+r)^{L-1}-1\bigr)G_{L-1}\|\B{X}\|_F.
\end{equation}
\end{proposition}

\begin{proof}
Since $\sigma_l(\B{0})=\B{0}$ and $\sigma_l$ are $\gamma_l$-Lipschitz,
\begin{equation}
    \|\B{X}_{l+1}\|_F=\|\sigma_l(\B{W}_l\B{X}_l)-\sigma_l(\B{0})\|_F
    \le \gamma_l\|\B{W}_l\B{X}_l\|_F
    \le \gamma_l\|\B{W}_l\|_2\|\B{X}_l\|_F.
\end{equation}
By induction,
\begin{equation}
    \label{eq:Xl_bound_rig}
    \|\B{X}_l\|_F \le \left(\prod_{t=1}^{l-1}\gamma_t\|\B{W}_t\|_2\right)\|\B{X}\|_F
    =G_{l-1}\|\B{X}\|_F,~~ l\ge1.
\end{equation}

Using Lipschitz continuity again,
\begin{align}
    \|\B{\delta}_{l+1}\|_F
    &=\|\sigma_l(\B{W}_l\B{X}_l)-\sigma_l(\widehat{\B{W}}_l\widehat{\B{X}}_l)\|_F\\
    &\le \gamma_l\|\B{W}_l\B{X}_l-\widehat{\B{W}}_l\widehat{\B{X}}_l\|_F.
\end{align}
Since $\widehat{\B{W}}_l=\B{W}_l+\B{E}_l$ and $\B{\delta}_l=\B{X}_l-\widehat{\B{X}}_l$,
\begin{equation}
    \B{W}_l\B{X}_l-\widehat{\B{W}}_l\widehat{\B{X}}_l
    =\B{W}_l\B{X}_l-(\B{W}_l+\B{E}_l)\widehat{\B{X}}_l
    =-\B{E}_l\B{X}_l+\widehat{\B{W}}_l\B{\delta}_l,
\end{equation}
hence, by the triangle inequality and $\|AB\|_F\le\|A\|_2\|B\|_F$,
\begin{equation}
    \label{eq:delta_rec_rig}
    \|\B{\delta}_{l+1}\|_F
    \le \gamma_l\Bigl(\|\B{E}_l\|_2\|\B{X}_l\|_F+\|\widehat{\B{W}}_l\|_2\|\B{\delta}_l\|_F\Bigr).
\end{equation}

By definition of $r$, $\|\B{E}_l\|_2\le r\|\B{W}_l\|_2$.
Additionally, $\|\widehat{\B{W}}_l\|_2\le\|\B{W}_l\|_2+\|\B{E}_l\|_2\le(1+r)\|\B{W}_l\|_2$.
Combining these with Eq.~\eqref{eq:Xl_bound_rig} in Eq.~\eqref{eq:delta_rec_rig} yields
\begin{equation}
    \|\B{\delta}_{l+1}\|_F
    \le \gamma_l\|\B{W}_l\|_2\Bigl(rG_{l-1}\|\B{X}\|_F+(1+r)\|\B{\delta}_l\|_F\Bigr).
\end{equation}
Define the normalized quantity
\begin{equation}
    a_l\coloneq\frac{\|\B{\delta}_l\|_F}{G_{l-1}\|\B{X}\|_F}, ~~l\ge1.
\end{equation}
Note $a_1=0$ because $\B{\delta}_1=\B{0}$.
Dividing the previous inequality by $G_l\|\B{X}\|_F$
(where $G_l=G_{l-1}\gamma_l\|\B{W}_l\|_2$) yields
\begin{equation}
    a_{l+1}\le r+(1+r)a_l.
\end{equation}
Let $b_l\coloneq a_l+1$. Then $b_{l+1}\le(1+r)b_l$ and $b_1=1$; hence
$b_l\le(1+r)^{l-1}$; therefore
\begin{equation}
    a_l\le(1+r)^{l-1}-1.
\end{equation}
Taking $l=L$ gives Eq.~\eqref{eq:uniform_bound_exact}.
\end{proof}

\begin{proposition}
\label{prop:exponential-error-upper}
Assume each activation $\sigma_l:\mab{R}^{n_l\times m}\to\mab{R}^{n_l\times m}$ is $\gamma_l$-Lipschitz with respect to the Frobenius norm and satisfies $\sigma_l(\B{0})=\B{0}$:
\begin{equation}
    \|\sigma_l(\B{U})-\sigma_l(\B{V})\|_F\le\gamma_l\|\B{U}-\B{V}\|_F.
\end{equation}
Assume moreover that $\sigma_l$ is Fr\'echet differentiable at the full-precision
pre-activation $\B{Y}_{l}\coloneq\B{W}_l\B{X}_l$ for each $l=1,\dots,L-1$. Let $\B{X}_1=\widehat{\B{X}}_1=\B{X}$ and for $l=1,\dots,L-1$ define
\begin{equation}
    \B{X}_{l+1}=\sigma_l(\B{W}_l\B{X}_l),~~
    \widehat{\B{X}}_{l+1}=\sigma_l(\widehat{\B{W}}_l \widehat{\B{X}}_l),~~
    \widehat{\B{W}}_l=\B{W}_l+\B{E}_l.
\end{equation}
Assume $\|\B{W}_k\|_2>0$ for $k=1,\dots,L-1$ and define
\begin{equation}
\B{G}_{L-1}\coloneq\prod_{l=1}^{L-1}\gamma_l\|\B{W}_l\|_2,~~
r\coloneq\max_{1\le k\le L-1}\frac{\|\B{E}_k\|_2}{\|\B{W}_k\|_2}.
\end{equation}

Define the \emph{first-order component} $\B{\delta}_L^{(1)}$ as follows.
For $t\in\mab{R}$ let $\widehat{\B{W}}_l(t)\coloneq\B{W}_l+t\B{E}_l$ and define
$\widehat{\B{X}}_1(t)\coloneq\B{X}$,
$\widehat{\B{X}}_{l+1}(t)\coloneq\sigma_l(\widehat{\B{W}}_l(t)\widehat{\B{X}}_l(t))$.
Let $\B{\delta}_l(t)\coloneq\B{X}_l-\widehat{\B{X}}_l(t)$.
Then $\B{\delta}_L^{(1)}$ is defined by the derivative
\begin{equation}
\B{\delta}_L^{(1)}\coloneq\left.\frac{d}{dt}\right|_{t=0}\B{\delta}_L(t).
\end{equation}
Under these assumptions,
\begin{equation}
\label{eq:delta-upper-general}
\|\B{\delta}_L^{(1)}\|_F
\le (L-1) r \B{G}_{L-1} \|\B{X}\|_F.
\end{equation}
In particular, if $\gamma_l\|\B{W}_l\|_2\le 1+\varepsilon$ for all $l$, then
\begin{equation}
\label{eq:delta-upper-exp}
\|\B{\delta}_L^{(1)}\|_F
\le (L-1) r (1+\varepsilon)^{L-1}\|\B{X}\|_F.
\end{equation}
\end{proposition}

\begin{proof}
Since $\sigma_l(\B{0})=\B{0}$ and $\sigma_l$ is $\gamma_l$-Lipschitz,
\begin{equation}
    \|\B{X}_{l+1}\|_F=\|\sigma_l(\B{W}_l\B{X}_l)-\sigma_l(\B{0})\|_F
    \le \gamma_l\|\B{W}_l\B{X}_l\|_F
    \le \gamma_l\|\B{W}_l\|_2\|\B{X}_l\|_F.
\end{equation}
By induction,
\begin{equation}
    \label{eq:Xl_bound}
    \|\B{X}_l\|_F \le \left(\prod_{t=1}^{l-1}\gamma_t\|\B{W}_t\|_2\right)\|\B{X}\|_F,~~ l\ge1.
\end{equation}

Fix $l\in\{1,\dots,L-1\}$.
By assumption, $\sigma_l$ is Fr\'echet differentiable at $\B{Y}_l\coloneq\B{W}_l\B{X}_l$.
Let $\B{J}_l\coloneq D\sigma_l(\B{Y}_l)$ denote its Fr\'echet derivative.
Because $\sigma_l$ is $\gamma_l$-Lipschitz, the operator norm of $\B{J}_l$, induced by the Frobenius norm, satisfies
\begin{equation}
    \label{eq:J_bound}
    \|\B{J}_l\|_{\mathrm{op}} \le \gamma_l.
\end{equation}
Indeed, for any $\B{H}$,
\begin{equation}
    \|\B{J}_l[\B{H}]\|_F
    =\lim_{t\to0}\frac{\|\sigma_l(\B{Y}_l+t\B{H})-\sigma_l(\B{Y}_l)\|_F}{|t|}
    \le \lim_{t\to0}\frac{\gamma_l\|t\B{H}\|_F}{|t|}=\gamma_l\|\B{H}\|_F.
\end{equation}

Now consider $\widehat{\B{X}}_{l+1}(t)=\sigma_l(\widehat{\B{W}}_l(t)\widehat{\B{X}}_l(t))$.
Since matrix multiplication is smooth and $\sigma_l$ is differentiable at $\B{Y}_l$,
the chain rule gives
\begin{equation}
    \left.\frac{d}{dt}\right|_{t=0}\widehat{\B{X}}_{l+1}(t)
    =\B{J}_l\!\left[\B{E}_l\B{X}_l+\B{W}_l\left.\frac{d}{dt}\right|_{t=0}\widehat{\B{X}}_l(t)\right].
\end{equation}
Because $\B{\delta}_l(t)=\B{X}_l-\widehat{\B{X}}_l(t)$, we have
$\left.\frac{d}{dt}\right|_{t=0}\widehat{\B{X}}_l(t)=-\B{\delta}_l^{(1)}$.
Therefore,
\begin{equation}
    \label{eq:first_order_recursion}
    \B{\delta}_{l+1}^{(1)}
    =-\B{J}_l[\B{E}_l\B{X}_l-\B{W}_l\B{\delta}_l^{(1)}]
    =-\B{J}_l(\B{E}_l\B{X}_l)+\B{J}_l(\B{W}_l\B{\delta}_l^{(1)}),
    ~~
    \B{\delta}_1^{(1)}=\B{0}.
\end{equation}

Taking Frobenius norms, using Eq.~\eqref{eq:J_bound} and $\| \B{A}\B{B}\|_F\le\|\B{A}\|_2\|\B{B}\|_F$,
\begin{equation}
    \label{eq:delta1_ineq}
    \|\B{\delta}_{l+1}^{(1)}\|_F
    \le \gamma_l\|\B{E}_l\|_2\|\B{X}_l\|_F
    +\gamma_l\|\B{W}_l\|_2\|\B{\delta}_l^{(1)}\|_F.
\end{equation}

Define $a_l\coloneq\|\B{\delta}_l^{(1)}\|_F$.
From Eq.~\eqref{eq:delta1_ineq} and $a_1=0$, straightforward induction yields
\begin{equation}
    a_L
    \le
    \sum_{k=1}^{L-1}
    \left(\prod_{s=k+1}^{L-1}\gamma_s\|\B{W}_s\|_2\right)\gamma_k\|\B{E}_k\|_2\|\B{X}_k\|_F.
\end{equation}
Apply Eq.~\eqref{eq:Xl_bound} to $\|\B{X}_k\|_F$ and factor out $\B{G}_{L-1}$:
\begin{equation}
    \|\B{\delta}_L^{(1)}\|_F
    \le
    \B{G}_{L-1}\left(\sum_{k=1}^{L-1}\frac{\|\B{E}_k\|_2}{\|\B{W}_k\|_2}\right)\|\B{X}\|_F
    \le
    (L-1)r\B{G}_{L-1}\|\B{X}\|_F,
\end{equation}
which is Eq.~\eqref{eq:delta-upper-general}.
If additionally $\gamma_l\|\B{W}_l\|_2\le1+\varepsilon$ for all $l$, then
$\B{G}_{L-1}\le(1+\varepsilon)^{L-1}$, proving Eq.~\eqref{eq:delta-upper-exp}.
\end{proof}

\begin{proposition}
\label{prop:exponential-lower-attainable}
Consider the 1-dimensional network ($d_l=n_l=1$) with $\sigma_l(z)=z$.
Let $W_l = 1+\varepsilon$ for all $l=1,\dots,L-1$ with $\varepsilon>0$,
and let the input be $X=C>0$.
Choose quantized weights
\begin{equation}
\widehat{W}_l = W_l + E_l,~~ E_l \equiv c_E>0,~~ l=1,\dots,L-1.
\end{equation}
Then, for all $L\ge2$, the exact activation mismatch at layer $L$ satisfies
\begin{equation}
\label{eq:scalar_lower}
    |\delta_L|
    = |X_L-\widehat{X}_L|
    \ge (L-1)c_EC(1+\varepsilon)^{L-2}.
\end{equation}
In particular,
\begin{equation}
\label{eq:scalar_lower_simplified}
    |\delta_L|
    \ge \frac{c_EC}{1+\varepsilon}(1+\varepsilon)^{L-1}.
\end{equation}
\end{proposition}

\begin{proof}
Since $\sigma_l$ is the identity map, we have
\begin{equation}
    X_L=(1+\varepsilon)^{L-1}C,~~
    \widehat{X}_L=(1+\varepsilon+c_E)^{L-1}C.
\end{equation}
Hence
\begin{equation}
    |\delta_L|
    =C\left|(1+\varepsilon+c_E)^{L-1}-(1+\varepsilon)^{L-1}\right|.
\end{equation}
Apply the mean value theorem to $f(t)=t^{L-1}$ on the interval
$[1+\varepsilon,\;1+\varepsilon+c_E]$:
there exists $\xi\in(1+\varepsilon,1+\varepsilon+c_E)$ such that
\begin{equation}
    (1+\varepsilon+c_E)^{L-1}-(1+\varepsilon)^{L-1}
    =f^{\prime}(\xi)c_E=(L-1)\xi^{L-2}c_E.
\end{equation}
Since $\xi\ge 1+\varepsilon$, we obtain
\begin{equation}
    |\delta_L|\ge (L-1)c_E(1+\varepsilon)^{L-2}C,
\end{equation}
which proves Eq.~\eqref{eq:scalar_lower}. Finally,
Eq.~\eqref{eq:scalar_lower_simplified} follows from
$(1+\varepsilon)^{L-2}=\nicefrac{1}{1+\varepsilon}(1+\varepsilon)^{L-1}$ and $L-1\ge1$ for $L\ge2$.
\end{proof}

\subsection{Derivation of Theorem~\ref{theorem:gurantee-reduction} and Corollary~\ref{cor:alpha-monotonicity}}\label{subsec:gurantee-reduce}

This section presents a rigorous statement and proof of Theorem~\ref{theorem:gurantee-reduction} and Corollary~\ref{cor:alpha-monotonicity}. 
We first formally restate Theorem~\ref{theorem:gurantee-reduction} below.
\begin{theorem}
\label{app-thm:main}
Consider an $L$-layer network
\begin{equation}
    \B{X}_{1}=\B{X},~~
    \B{X}_{l+1}=\sigma_l(\B{W}_l\B{X}_l),~~ l=1,\dots,L.
\end{equation}
Assume each $\sigma_l$ is $\gamma_l$-Lipschitz with respect to $\|\cdot\|_F$ and satisfies $\sigma_l(\B{0})=\B{0}$.
Let the quantized forward pass be
\begin{equation}
    \widehat{\B{X}}_{1}=\B{X},~~
    \widehat{\B{X}}_{l+1}=\sigma_l(\widehat{\B{W}}_l\widehat{\B{X}}_l).
\end{equation}
Define the activation mismatch as $\B{\delta}_l \coloneq \B{X}_l-\widehat{\B{X}}_l$.

Fix any matrices $\B{E}_l$ and set
\begin{equation}
    \widehat{\B{W}}_l^{\mathrm{BASE}}\coloneq\B{W}_l+\B{E}_l.
\end{equation}
For QEP, define each $l$
\begin{equation}
    \B{W}_l^\ast(\alpha_l)
    \coloneq\B{W}_l+\alpha_l\B{W}_l\B{\delta}_l\widehat{\B{X}}_l^\top\widehat{\B{H}}_l^{-1},
    ~~ \alpha_l\in[0,1],
\end{equation}
and set
\begin{equation}
    \widehat{\B{W}}_l^{\mathrm{QEP}}\coloneq\B{W}_l^\ast(\alpha_l)+\B{E}_l.
\end{equation}

Define the per-layer pre-activation residuals
\begin{equation}
    \B{R}_l^{M}\coloneq\B{W}_l\B{X}_l-\widehat{\B{W}}_l^{M}\widehat{\B{X}}_l^{M},
    ~~ M\in\{\mathrm{BASE},\mathrm{QEP}\},
\end{equation}
 the global Lipschitz upper bound on the output mismatch
\begin{equation}
    \mathcal{U}^{M}\coloneq\sum_{k=1}^{L}
    \left(\prod_{s=k+1}^{L}\gamma_s\|\B{W}_s\|_2\right)\gamma_k\|\B{R}_k^{M}\|_F.
\end{equation}
Then for every choice of $\{\alpha_l\}_{l=1}^L\subset[0,1]$,
\begin{equation}
    \mathcal{U}^{\mathrm{QEP}}\le \mathcal{U}^{\mathrm{BASE}}.
\end{equation}
Consequently,
\begin{equation}
    \|\B{\delta}_{L+1}^{\mathrm{QEP}}\|_F \le \mathcal{U}^{\mathrm{QEP}}
    \le \mathcal{U}^{\mathrm{BASE}},
    ~~
    \|\B{\delta}_{L+1}^{\mathrm{BASE}}\|_F \le \mathcal{U}^{\mathrm{BASE}}.
\end{equation}
\end{theorem}

\begin{proof}
Fix a method $M\in\{\mathrm{BASE},\mathrm{QEP}\}$.
By Lipschitz continuity,
\begin{equation}
    \|\B{\delta}_{l+1}^{M}\|_F
    =\|\sigma_l(\B{W}_l\B{X}_l)-\sigma_l(\widehat{\B{W}}_l^{M}\widehat{\B{X}}_l^{M})\|_F
    \le \gamma_l\|\B{W}_l\B{X}_l-\widehat{\B{W}}_l^{M}\widehat{\B{X}}_l^{M}\|_F
    =\gamma_l\|\B{R}_l^{M}\|_F.
\end{equation}
Iterating this inequality through the remaining layers yields
\begin{equation}
    \|\B{\delta}_{L+1}^{M}\|_F
    \le
    \sum_{k=1}^{L}
    \left(\prod_{s=k+1}^{L}\gamma_s\|\B{W}_s\|_2\right)\gamma_k\|\B{R}_k^{M}\|_F
    =\mathcal{U}^{M}.
\end{equation}

It remains to show $\mathcal{U}^{\mathrm{QEP}}\le\mathcal{U}^{\mathrm{BASE}}$.
We prove a stronger per-layer inequality:
\begin{equation}
\label{eq:layer_residual_compare}
    \|\B{R}_l^{\mathrm{QEP}}\|_F \le \|\B{R}_l^{\mathrm{BASE}}\|_F,~~\forall l.
\end{equation}
Fix $l$ and write $\widehat{\B{X}}_l\coloneq\widehat{\B{X}}_l^{\mathrm{QEP}}$ and $\B{\delta}_l\coloneq\B{X}_l-\widehat{\B{X}}_l$.
Define the orthogonal projection
\begin{equation}
    \B{P}_l\coloneq\widehat{\B{X}}_l^\top(\widehat{\B{X}}_l\widehat{\B{X}}_l^\top)^{-1}\widehat{\B{X}}_l,
\end{equation}
which satisfies $\B{P}_l^2=\B{P}_l$ and $\B{P}_l^\top=\B{P}_l$.

By the construction of $\B{W}_l^\ast(\alpha_l)$, we have the exact identity
\begin{equation}
    \B{W}_l^\ast(\alpha_l)\widehat{\B{X}}_l
    =
    \B{W}_l\widehat{\B{X}}_l+\alpha_l\B{W}_l\B{\delta}_l\B{P}_l.
\end{equation}
Therefore, using $\widehat{\B{W}}_l^{\mathrm{QEP}}=\B{W}_l^\ast(\alpha_l)+\B{E}_l$,
\begin{align}
    \B{R}_l^{\mathrm{QEP}}
    &=\B{W}_l\B{X}_l-(\B{W}_l^\ast(\alpha_l)+\B{E}_l)\widehat{\B{X}}_l\\
    &=\B{W}_l(\widehat{\B{X}}_l+\B{\delta}_l)-\B{W}_l^\ast(\alpha_l)\widehat{\B{X}}_l-\B{E}_l\widehat{\B{X}}_l\\
    &=\B{W}_l\B{\delta}_l-\alpha_l\B{W}_l\B{\delta}_l\B{P}_l-\B{E}_l\widehat{\B{X}}_l\\
    &=\B{W}_l\B{\delta}_l(\B{I}-\alpha_l\B{P}_l)-\B{E}_l\widehat{\B{X}}_l.
\end{align}

Consider the BASE residual at the same layer evaluated on the same $\widehat{\B{X}}_l$:
\begin{equation}
    \widetilde{\B{R}}_l^{\mathrm{BASE}}
    \coloneq\B{W}_l\B{X}_l-(\B{W}_l+\B{E}_l)\widehat{\B{X}}_l
    =\B{W}_l\B{\delta}_l-\B{E}_l\widehat{\B{X}}_l.
\end{equation}
Hence
\begin{equation}
    \B{R}_l^{\mathrm{QEP}}
    =
    \widetilde{\B{R}}_l^{\mathrm{BASE}}-\alpha_l\B{W}_l\B{\delta}_l\B{P}_l.
\end{equation}
Since $\B{P}_l$ is an orthogonal projection and $0\le \alpha_l\le 1$,
Lemma~\ref{lem:mono_projection} implies
\begin{equation}
    \|\B{W}_l\B{\delta}_l(\B{I}-\alpha_l\B{P}_l)\|_F
    \le
    \|\B{W}_l\B{\delta}_l\|_F.
\end{equation}
Therefore, by the triangle inequality,
\begin{equation}
    \|\B{R}_l^{\mathrm{QEP}}\|_F
    =\|\B{W}_l\B{\delta}_l(\B{I}-\alpha_l\B{P}_l)-\B{E}_l\widehat{\B{X}}_l\|_F
    \le 
    \|\widetilde{\B{R}}_l^{\mathrm{BASE}}\|_F.
\end{equation}
Finally, since $\widehat{\B{X}}_l^{\mathrm{BASE}}$ is the BASE activation produced by the BASE recursion,
$\B{R}_l^{\mathrm{BASE}}$ is exactly $\widetilde{\B{R}}_l^{\mathrm{BASE}}$ evaluated at $\widehat{\B{X}}_l^{\mathrm{BASE}}$.
Thus, taking $\widehat{\B{X}}_l=\widehat{\B{X}}_l^{\mathrm{BASE}}$ in the above inequality yields
Eq.~\eqref{eq:layer_residual_compare}. Summing with nonnegative weights
$(\prod_{s=k+1}^{L}\gamma_s\|\B{W}_s\|_2)\gamma_k$ yields
$\mathcal{U}^{\mathrm{QEP}}\le \mathcal{U}^{\mathrm{BASE}}$, thus completing the proof.
\end{proof}

We further demonstrate that the final quantization error decreases monotonically as each propagation strength parameter $\alpha_l$ approaches $1$. 

\begin{corollary}
\label{prop:monotonicity-alpha}
Fix the layer index $l\in\{1,\dots,L\}$ and assume
that $\widehat{\B{H}}_l\coloneq\widehat{\B{X}}_l\widehat{\B{X}}_l^\top$ is invertible.
Let the activation mismatch be $\B{\delta}_l\coloneq\B{X}_l-\widehat{\B{X}}_l$.
Define the orthogonal projection
\begin{equation}
    \B{P}_l
    \coloneq
    \widehat{\B{X}}_l^\top(\widehat{\B{X}}_l\widehat{\B{X}}_l^\top)^{-1}\widehat{\B{X}}_l
    \in\mab{R}^{m\times m}.
\end{equation}
For $\alpha\in[0,1]$, define the QEP corrected weight in the continuous domain as
\begin{equation}
    \B{W}_l^\ast(\alpha)
    \coloneq
    \B{W}_l+\alpha\B{W}_l\B{\delta}_l\widehat{\B{X}}_l^\top\widehat{\B{H}}_l^{-1}.
\end{equation}
 Let $\widehat{\B{W}}_l(\alpha)\coloneq\B{W}_l^\ast(\alpha)+\B{E}_l$ be defined by a fixed matrix $\B{E}_l$.
Then, for any $0\le \alpha^{\prime}\le \alpha\le 1$,
\begin{equation}
    \label{eq:mono-propagation-term}
    \|\B{W}_l\B{\delta}_l(\B{I}-\alpha\B{P}_l)\|_F
    \le
    \|\B{W}_l\B{\delta}_l(\B{I}-\alpha^{\prime}\B{P}_l)\|_F.
\end{equation}
Moreover, the pre-activation residual satisfies the exact identity
\begin{equation}
    \label{eq:exact-residual-alpha}
    \B{W}_l\B{X}_l-\widehat{\B{W}}_l(\alpha)\widehat{\B{X}}_l
    =
    \B{W}_l\B{\delta}_l(\B{I}-\alpha\B{P}_l)-\B{E}_l\widehat{\B{X}}_l,
\end{equation}
and hence the following upper bound is also monotone in $\alpha$:
\begin{equation}
    \label{eq:mono-upperbound-residual}
    \|\B{W}_l\B{X}_l-\widehat{\B{W}}_l(\alpha)\widehat{\B{X}}_l\|_F
    \le
    \|\B{W}_l\B{\delta}_l(\B{I}-\alpha\B{P}_l)\|_F+\|\B{E}_l\|_2\|\widehat{\B{X}}_l\|_F.
\end{equation}
\end{corollary}

\begin{proof}
First, $\B{P}_l$ is an orthogonal projection. Indeed,
\begin{equation}
\B{P}_l^\top
=
\widehat{\B{X}}_l^\top(\widehat{\B{X}}_l\widehat{\B{X}}_l^\top)^{-1}\widehat{\B{X}}_l
=\B{P}_l,
\end{equation}
and
\begin{equation}
\B{P}_l^2
=
\widehat{\B{X}}_l^\top(\widehat{\B{X}}_l\widehat{\B{X}}_l^\top)^{-1}
\underbrace{\widehat{\B{X}}_l\widehat{\B{X}}_l^\top}_{=\widehat{\B{H}}_l}
(\widehat{\B{X}}_l\widehat{\B{X}}_l^\top)^{-1}\widehat{\B{X}}_l
=\B{P}_l.
\end{equation}
Thus, Lemma~\ref{lem:mono_projection} applies with
$\B{Z}\coloneq\B{W}_l\B{\delta}_l$ and $\B{P}\coloneq\B{P}_l$, yielding
\begin{equation}
    \|\B{W}_l\B{\delta}_l(\B{I}-\alpha\B{P}_l)\|_F
    \le
    \|\B{W}_l\B{\delta}_l(\B{I}-\alpha^{\prime}\B{P}_l)\|_F,
\end{equation}
which proves Eq.~\eqref{eq:mono-propagation-term}.

Next, using the definitions of $\B{W}_l^\ast(\alpha)$ and $\widehat{\B{H}}_l^{-1}=(\widehat{\B{X}}_l\widehat{\B{X}}_l^\top)^{-1}$,
\begin{equation}
    \B{W}_l^\ast(\alpha)\widehat{\B{X}}_l
    =
    \B{W}_l\widehat{\B{X}}_l
    +\alpha\B{W}_l\B{\delta}_l\widehat{\B{X}}_l^\top(\widehat{\B{X}}_l\widehat{\B{X}}_l^\top)^{-1}\widehat{\B{X}}_l
    =
    \B{W}_l\widehat{\B{X}}_l+\alpha\B{W}_l\B{\delta}_l\B{P}_l.
    \end{equation}
    Therefore, since $\widehat{\B{W}}_l(\alpha)=\B{W}_l^\ast(\alpha)+\B{E}_l$ and $\B{X}_l=\widehat{\B{X}}_l+\B{\delta}_l$,
\begin{align}
    \B{W}_l\B{X}_l-\widehat{\B{W}}_l(\alpha)\widehat{\B{X}}_l
    &=\B{W}_l(\widehat{\B{X}}_l+\B{\delta}_l)-(\B{W}_l^\ast(\alpha)+\B{E}_l)\widehat{\B{X}}_l\\
    &=\B{W}_l\B{\delta}_l-\alpha\B{W}_l\B{\delta}_l\B{P}_l-\B{E}_l\widehat{\B{X}}_l\\
    &=\B{W}_l\B{\delta}_l(\B{I}-\alpha\B{P}_l)-\B{E}_l\widehat{\B{X}}_l,
\end{align}
which is Eq.~\eqref{eq:exact-residual-alpha}.
Finally, Eq.~\eqref{eq:mono-upperbound-residual} follows from the triangle inequality and
$\|\B{E}_l\widehat{\B{X}}_l\|_F\le \|\B{E}_l\|_2\|\widehat{\B{X}}_l\|_F$.
\end{proof}

\subsection{Relationship of QEP Correction and Ridge Regularization}\label{sec:alpha_lambda_equiv}

We formally establish a rigorous mathematical connection between the Quantization Error Propagation (QEP) correction parameter $\alpha_l$ and the ridge regularization parameter $\lambda$. We show that tuning the QEP parameter $\alpha_l$ is equivalent to adjusting the strength of ridge regularization with parameter $\lambda$. We prove the monotone inverse relationship between these two parameters.

\begin{proposition}\label{app-prop:alpha_lambda_correspondence}
    The QEP update with mixing factor $\alpha_{l} \in [0, 1]$ is
    \begin{equation}
        \widehat{\B{W}}_{l}^{\ast}(\alpha_{l}) = \B{W}_{l} (\B{I} + \alpha \B{\delta}_{l} \widehat{\B{X}}_{l}^{\top} \widehat{\B{H}}_{l}^{-1}).
    \end{equation}
    The unique minimizer of the ridge objective
    \begin{equation}
        \min_{\widehat{\B{W}}_{l} \in \mab{R}^{n_{l} \times d_{l}}} f(\widehat{\B{W}}_{l}),~~f(\widehat{\B{W}}_{l}) = \|\B{W}_{l} \B{X}_{l} - \widehat{\B{W}}_{l} \widehat{\B{X}}_{l} \|_{F}^{2} + \lambda_{l} \|\B{W}_{l} - \widehat{\B{W}}_{l}\|_{F}^{2},~~\lambda_{l} \ge 0,
    \end{equation}
    equals
    \begin{equation}
    \label{eq:optimal-lambda-quant-w}
        \widehat{\B{W}}^{\ast}_{l}(\lambda_{l}) = \B{W}_{l}\left(\B{I} + \B{\delta}_{l} \widehat{\B{X}}_{l}^{\top} (\widehat{\B{H}}_{l} + \lambda \B{I} )^{-1} \right).
    \end{equation}
    Let the positive definite matrices be 
    \begin{equation}
        \B{G}(\alpha_{l}) \coloneqq \alpha \hat{\B{H}}_{l}^{-1},~~~\B{R}(\lambda_{l}) \coloneqq (\widehat{\B{H}}_{l} + \lambda \B{I})^{-1}.
    \end{equation}
    Then 
    \begin{equation}
        \alpha_{1} \le \alpha_{2} \Rightarrow \B{G}(\alpha_{1}) \preceq \B{G}(\alpha_{2}),~~\lambda_{1} \le \lambda_{2} \Rightarrow \B{R}(\lambda_{1}) \succeq \B{R}(\lambda_{2}),
    \end{equation}
    and the scalar mapping as follows:
    \begin{equation}
        \alpha(\lambda) \coloneqq \frac{1}{d} \mathrm{Tr} \widehat{\B{H}}_{l} \B{R}(\lambda) = \frac{1}{d_{l}} \sum_{i=1}^{d_{l}} \frac{\gamma_{i}}{\gamma_{i} + \lambda_{l}}
    \end{equation}
    with $\gamma_{1} \ge \cdots \ge \gamma_{d_{1}} > 0$ the eigenvalues of $\widehat{\B{H}}_{l}$, is strictly decreasing, satisfies $\alpha(0) = 1$ and $\lim_{\lambda \to \infty} \alpha(\lambda) = 0$, and obeys
    \begin{equation}
        \mathrm{Tr} \widehat{\B{H}}_{l} \B{G}(\alpha(\lambda)) = \mathrm{Tr} \widehat{\B{H}}_{l} \B{R}(\lambda).
    \end{equation}
    Thus, decreasing $\lambda$ from $+ \infty$ to $0$ corresponds to increasing $\alpha_{l}$ from $0$ to $1$.
\end{proposition}

\begin{proof}

A standard differential identity $\partial \|\B{A}\|_{F}^{2} = 2\B{A}$ gives
\begin{equation}
    \nabla_{\widehat{\B{W}}_{l}} f(\widehat{\B{W}}_{l}) = 2 \left(\widehat{\B{W}}_{l} \widehat{\B{H}}_{l} - \B{W}_{l} \B{X}_{l} \widehat{\B{X}}_{l}^{\top} \right) + 2\lambda (\widehat{\B{W}}_{l} - \B{W}_{l}).
\end{equation}
Setting this gradient $\B{0}$ yields
\begin{equation}
    \widehat{\B{W}}_{l}(\widehat{\B{H}}_{l} + \lambda \B{I}) = \B{W}_{l} (\B{X}_{l} \widehat{\B{X}}_{l}^{\top} + \lambda \B{I}),
\end{equation}
and right multiplication by inverse of $\widehat{\B{H}}_{l} + \lambda \B{I}$ produces Eq.~\eqref{eq:optimal-lambda-quant-w}. 
Convexity of $f$ ensures uniqueness.

Diagonalise $\widehat{\B{H}}_{l} = \B{U} \B{\Gamma} \B{U}^{\top}$ with $\Gamma = \mathrm{diag}(\gamma_{1}, \ldots, \gamma_{d_{l}})$. 
Then $\B{G}(\alpha_{l}) = \B{U} \alpha_{l} \B{\Gamma}^{-1} \B{U}^{\top}$ has eigenvalues $\nicefrac{\alpha_{l}}{\gamma_{l}}$, which increase strictly with $\alpha_{l}$, while $\B{R}(\lambda_{l}) = \B{U}(\B{\Gamma}+\lambda \B{I})^{-1} \B{U}^{\top}$ has eigenvalues $\nicefrac{1}{\gamma_{i}+\lambda}$, which decrease with $\lambda_{l}$, which means Loewner relations follow directly.

Furthermore, the following equation holds:
\begin{equation}
    \alpha(\lambda) = \frac{1}{d_{1}} \mathrm{Tr}(\widehat{\B{H}}_{l} \B{R}(\lambda)) = \frac{1}{d_{l}} \sum_{i=1}^{d_{1}} \frac{\gamma_{i}}{\gamma_{i} + \lambda}
\end{equation}
Each summand has derivative
\begin{equation}
    \frac{\partial}{\partial \lambda} \frac{\gamma_{i}}{\gamma_{i} + \lambda} = - \frac{\gamma_{i}}{(\gamma_{i}+\lambda)^{2}} < 0,
\end{equation}
which means $\alpha^{\prime}(\lambda)<0, \forall \lambda \ge 0$. Thus, $\alpha(\cdot)$ is strictly decreasing on $[0, \infty)$. 
One has 
\begin{equation}
    \lim_{\lambda \to 0} \frac{\gamma_{i}}{\gamma_{i} + \lambda} = 1,~~~\lim_{\lambda \to \infty} \frac{\gamma_{i}}{\gamma_{i} + \lambda} = 0.
\end{equation}
$\alpha(\cdot)$ is strictly decreasing from $1$ to $0$ and smooth on $[0, \infty)$. 
Because $\alpha$ is continous, strictly decreasing, it is a bijection from $[0, +\infty)$ onto $(0, 1]$. By construction
\begin{equation}
    \mathrm{Tr} \widehat{\B{H}}_{l} \B{G}(\alpha(\lambda)) = \mathrm{Tr} \widehat{\B{H}}_{l} \B{R}(\lambda).
\end{equation}
Thus, decreasing $\lambda$ from $+ \infty$ to $0$ corresponds to increasing $\alpha_{l}$ from $0$ to $1$.

\end{proof}

\subsection{Technical Lemma}\label{subsec:technical-lemma}

\begin{lemma}
\label{lem:mono_projection}
Let $\B{Z}\in\mab{R}^{m \times n}$ be arbitrary, and let
$P\in \mab{R}^{n\times n}$ be an orthogonal projection, i.e., $\B{P}^{2}=\B{P}$ and $\B{P}^{\top}=\B{P}$.
For every pair $0 \le \alpha^{\prime} \le \alpha \le 1$,
\begin{equation}
  \|\B{Z}(\B{I}-\alpha \B{P})\|_{F}
  \le
  \|\B{Z}(I-\alpha^{\prime} \B{P})\|_{F}
  \le
  \|\B{Z}\|_{F}.
  \label{eq:projection-mono}
\end{equation}
\end{lemma}

\begin{proof}
Write $f(\alpha) \coloneqq \|\B{Z}(\B{I}-\alpha \B{P})\|_{F}^{2}$.
Because $\B{P}^{\top}=\B{P}$ and $\B{P}^{2}=\B{P}$,
\begin{equation}
  f(\alpha)
  =\mathrm{Tr} \left[(\B{I}-\alpha \B{P})\B{Z}^{\top}\B{Z}(\B{I}-\alpha \B{P})\right]
  =\|\B{Z}\|_{F}^{2}-2\alpha(1-\alpha)
    \underbrace{\mathrm{Tr}(\B{Z}^{\top}\B{Z}\B{P})}_{t\ge 0}.  
\end{equation}
Thus, $f^{\prime}(\alpha)=-(2-\alpha)t\le 0$ on $[0,1]$ indicates that $f(\alpha)$ is
non-increasing.  Taking square roots yields the first inequality in Eq.~\eqref{eq:projection-mono}.  
Setting $\alpha^{\prime}=0$ yields the second inequality:
$\|\B{Z}(\B{I}-\alpha \B{P})\|_{F}\le\|\B{Z}\|_{F}$.
\end{proof}

\section{Additional Implementation Details}\label{sec:add-implementation}

\subsection{Damping for Hessian}\label{subsec:damping}
A standard numerical issue in PTQ arises when the Hessian matrix $\widehat{\B{H}}_l$ is ill-conditioned or singular, rendering its inversion unstable or undefined.
Following GPTQ~\citep{frantar2022gptq}, we resolve this issue by employing a damping strategy that adds a small scalar value $\lambda$ to the diagonal elements of $\widehat{\B{H}}_l$ to ensure positive definiteness.
In our implementation, we set $\lambda$ to the mean of the diagonal elements of $\widehat{\B{H}}_l$, providing a straightforward yet effective method to stabilize the inversion process.

\begin{figure}[tb]
    \centering
    \includegraphics[width=\linewidth]{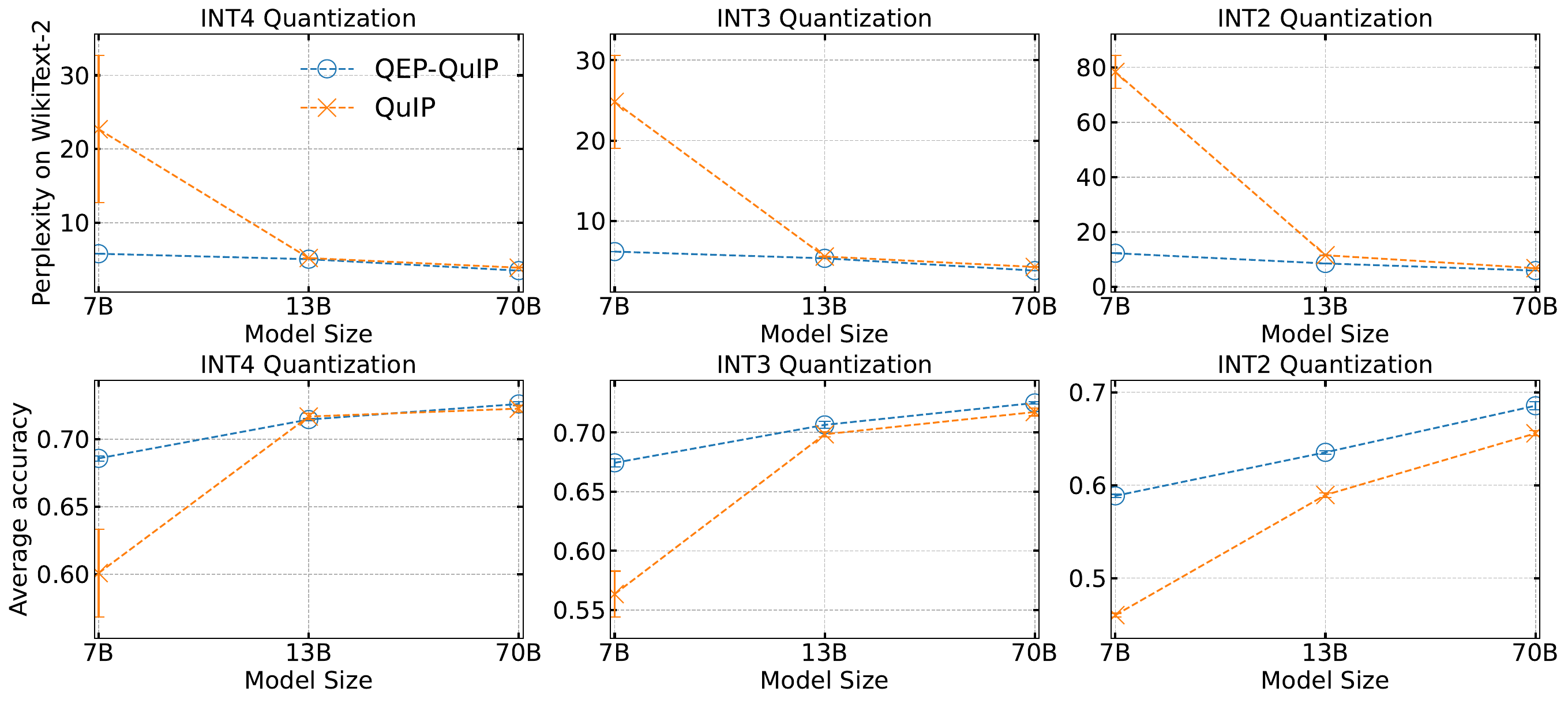}
    \caption{Results averaged over 5 random seeds comparing QuIP with and without QEP across different quantization levels.
    Each subplot shows results for INT4, INT3, and INT2 quantization, respectively, with the horizontal axis indicating model size (7B, 13B, 70B). The top row reports perplexity on WikiText-2 (lower is better), while the bottom row shows the average of normalized accuracy scores on ARC (easy), PIQA, and StoryCloze benchmarks (higher is better), representing generalization capability. Error bars represent the standard error of the mean (SEM). Models using QEP-QuIP consistently outperform or match the performance of baseline QuIP, especially under more aggressive quantization (INT3 and INT2).}
    \label{fig:quip-results}
\end{figure}

\section{Additional Experiments}\label{sec:add-experiments}

\subsection{Additional Perplexity Results}\label{app-subsec:add-perplexity-results}
Due to space constraints, the main text reports perplexity results solely for the WikiText-2 dataset. Here, we provide additional results for PTB (Table~\ref{tab:gqep_ptb_full}) and C4 (Table~\ref{tab:gqep_c4_full}), along with extended results for WikiText-2 (Table~\ref{tab:gqep_full_noquip}). These supplementary results further validate that QEP consistently enhances PTQ performance, particularly in low-bit quantization scenarios.

\begin{table}[tb]
  \centering
  \small
  \caption{Perplexities ($\downarrow$) on WikiText-2 for Llama-2 (7B, 13B, 70B) under eight quantization settings.}
  \label{tab:gqep_full_noquip}

  \begin{tabular}{c|cc|ccc}
    \toprule
    \textbf{Bits} & \textbf{Method} & \textbf{QEP} &
    \textbf{Llama-2-7B} & \textbf{Llama-2-13B} & \textbf{Llama-2-70B} \\
    \midrule
    \multirow{6}{*}{INT4g128}
      & \multirow{2}{*}{RTN}  & \xmark & 5.726 & 4.984 & 3.463 \\
      &                       & \cmark & \textbf{\q{5.687}} & \textbf{\q{4.966}} & \textbf{\q{3.431}} \\
      & \multirow{2}{*}{GPTQ} & \xmark & 5.698 & 4.987 & 3.419 \\
      &                       & \cmark & \textbf{\q{5.609}} & \textbf{\q{4.969}} & \textbf{\q{3.416}} \\
      & \multirow{2}{*}{AWQ}  & \xmark & 5.599 & 4.987 & 3.408 \\
      &                       & \cmark & \textbf{\q{5.580}} & \textbf{\q{4.969}} & \textbf{\q{3.404}} \\
    \midrule
    \multirow{6}{*}{INT4}
      & \multirow{2}{*}{RTN}  & \xmark & 6.116 & 5.206 & 3.672 \\
      &                       & \cmark & \textbf{\q{6.017}} & \textbf{\q{5.165}} & \textbf{\q{3.621}} \\
      & \multirow{2}{*}{GPTQ} & \xmark & 6.083 & 5.167 & 3.594 \\
      &                       & \cmark & \textbf{\q{5.933}} & \textbf{\q{5.127}} & \textbf{\q{3.576}} \\
      & \multirow{2}{*}{AWQ}  & \xmark & 5.831 & 5.064 & 3.484 \\
      &                       & \cmark & \textbf{\q{5.756}} & \textbf{\q{5.041}} & \textbf{\q{3.479}} \\
    \midrule
    \multirow{6}{*}{INT3g128}
      & \multirow{2}{*}{RTN}  & \xmark & 6.662 & 5.518 & 3.978 \\
      &                       & \cmark & \textbf{\q{6.330}} & \textbf{\q{5.412}} & \textbf{\q{3.882}} \\
      & \multirow{2}{*}{GPTQ} & \xmark & 6.411 & 5.459 & 3.880 \\
      &                       & \cmark & \textbf{\q{6.160}} & \textbf{\q{5.358}} & \textbf{\q{3.838}} \\
      & \multirow{2}{*}{AWQ}  & \xmark & 6.247 & 5.315 & 3.740 \\
      &                       & \cmark & \textbf{\q{6.108}} & \textbf{\q{5.295}} & \textbf{\q{3.724}} \\
    \midrule
    \multirow{6}{*}{INT3}
      & \multirow{2}{*}{RTN}  & \xmark & 539.866 & 10.688 & 7.530 \\
      &                       & \cmark & \textbf{\q{17.309}} & \textbf{\q{7.458}} & \textbf{\q{5.648}} \\
      & \multirow{2}{*}{GPTQ} & \xmark & 10.881 & 6.632 & 4.860 \\
      &                       & \cmark & \textbf{\q{7.898}} & \textbf{\q{6.245}} & \textbf{\q{4.102}} \\
      & \multirow{2}{*}{AWQ}  & \xmark & 15.299 & 6.448 & 4.362 \\
      &                       & \cmark & \textbf{\q{11.131}} & \textbf{\q{6.092}} & \textbf{\q{4.103}} \\
    \midrule
    \multirow{6}{*}{INT2g32}
      & \multirow{2}{*}{RTN}  & \xmark & 90.692 & 10.563 & 6.802 \\
      &                       & \cmark & \textbf{\q{12.249}} & \textbf{\q{7.920}} & \textbf{\q{5.869}} \\
      & \multirow{2}{*}{GPTQ} & \xmark & 12.023 & 8.394 & 5.621 \\
      &                       & \cmark & \textbf{\q{9.245}} & \textbf{\q{7.362}} & \textbf{\q{5.445}} \\
      & \multirow{2}{*}{AWQ}  & \xmark & 15887.204 & 106933.227 & 63663.707 \\
      &                       & \cmark & \textbf{\q{51.874}} & \textbf{\q{80654.797}} & \textbf{\q{37096.516}} \\
    \midrule
    \multirow{6}{*}{INT2g64}
      & \multirow{2}{*}{RTN}  & \xmark & 431.595 & 26.220 & 10.312 \\
      &                       & \cmark & \textbf{\q{19.371}} & \textbf{\q{9.917}} & \textbf{\q{6.992}} \\
      & \multirow{2}{*}{GPTQ} & \xmark & 278.302 & 11.584 & 6.546 \\
      &                       & \cmark & \textbf{\q{14.737}} & \textbf{\q{8.685}} & \textbf{\q{6.030}} \\
      & \multirow{2}{*}{AWQ}  & \xmark & \textbf{217111.860} & \textbf{121737.148} & \textbf{71703.781} \\
      &                       & \cmark & \q{241136.594} & \q{126944.578} & \q{74227.539} \\
    \midrule
    \multirow{6}{*}{INT2g128}
      & \multirow{2}{*}{RTN}  & \xmark & 4270.828 & 122.063 & 27.268 \\
      &                       & \cmark & \textbf{\q{35.291}} & \textbf{\q{12.779}} & \textbf{\q{8.799}} \\
      & \multirow{2}{*}{GPTQ} & \xmark & 43.915 & \textbf{16.653} & 8.123 \\
      &                       & \cmark & \textbf{\q{17.886}} & \q{19.952} & \textbf{\q{6.825}} \\
      & \multirow{2}{*}{AWQ}  & \xmark & \textbf{222344.250} & \textbf{122795.898} & \textbf{72446.680} \\
      &                       & \cmark & \q{247751.203} & \q{126813.172} & \q{74192.570} \\
    \midrule
    \multirow{6}{*}{INT2}
      & \multirow{2}{*}{RTN}  & \xmark & \textbf{17783.918} & \textbf{51152.832} & 26077.172 \\
      &                       & \cmark & \q{97153.266} & \q{61158.555} & \textbf{\q{26063.672}} \\
      & \multirow{2}{*}{GPTQ} & \xmark & 13051.469 & \textbf{1301.395} & 107.458 \\
      &                       & \cmark & \textbf{\q{7214.328}} & \q{2782.353} & \textbf{\q{52.472}} \\
      & \multirow{2}{*}{AWQ}  & \xmark & \textbf{199448.797} & 93036.517 & \textbf{81834.344} \\
      &                       & \cmark & \q{229888.406} & \textbf{\q{74735.836}} & \q{88684.156} \\
    \bottomrule
  \end{tabular}
\end{table}

\begin{table}[tb]
  \centering
  \small
  \caption{Perplexities ($\downarrow$) on PTB for Llama-2 (7B, 13B, 70B) under eight quantization settings. ``N/A'' denotes numerical overflow (NaN).}
  \label{tab:gqep_ptb_full}
  \setlength{\tabcolsep}{4pt}
  \renewcommand{\arraystretch}{1.12}

  \begin{tabular}{c|cc|ccc}
    \toprule
    \textbf{Bits} & \textbf{Method} & \textbf{QEP} &
    \textbf{Llama-2-7B} & \textbf{Llama-2-13B} & \textbf{Llama-2-70B} \\
    \midrule
    \multirow{6}{*}{INT4g128}
      & \multirow{2}{*}{RTN}  & \xmark & 61.750 & 53.835 & \textbf{24.146} \\
      &                       & \cmark & \textbf{\q{47.798}} & \textbf{\q{49.503}} & \q{24.604} \\
      & \multirow{2}{*}{GPTQ} & \xmark & N/A & 51.133 & \textbf{24.101} \\
      &                       & \cmark & \q{N/A} & \textbf{\q{50.072}} & \q{24.243} \\
      & \multirow{2}{*}{AWQ}  & \xmark & 43.894 & \textbf{53.863} & \textbf{24.525} \\
      &                       & \cmark & \textbf{\q{40.445}} & \q{55.345} & \q{24.554} \\
    \midrule
    \multirow{6}{*}{INT4}
      & \multirow{2}{*}{RTN}  & \xmark & 82.641 & 60.749 & 23.545 \\
      &                       & \cmark & \textbf{\q{50.168}} & \textbf{\q{53.117}} & \textbf{\q{23.346}} \\
      & \multirow{2}{*}{GPTQ} & \xmark & N/A & 53.561 & 24.720 \\
      &                       & \cmark & \textbf{\q{124291.961}} & \textbf{\q{53.537}} & \textbf{\q{24.149}} \\
      & \multirow{2}{*}{AWQ}  & \xmark & 60.261 & \textbf{56.152} & 25.542 \\
      &                       & \cmark & \textbf{\q{46.937}} & \q{57.445} & \textbf{\q{24.411}} \\
    \midrule
    \multirow{6}{*}{INT3g128}
      & \multirow{2}{*}{RTN}  & \xmark & 55.467 & 64.638 & \textbf{23.586} \\
      &                       & \cmark & \textbf{\q{48.576}} & \textbf{\q{54.866}} & \q{24.776} \\
      & \multirow{2}{*}{GPTQ} & \xmark & N/A & \textbf{57.079} & \textbf{24.091} \\
      &                       & \cmark & \q{N/A} & \q{62.083} & \q{24.092} \\
      & \multirow{2}{*}{AWQ}  & \xmark & 64.932 & \textbf{57.273} & \textbf{24.668} \\
      &                       & \cmark & \textbf{\q{52.356}} & \q{61.479} & \q{26.309} \\
    \midrule
    \multirow{6}{*}{INT3}
      & \multirow{2}{*}{RTN}  & \xmark & \textbf{37167.801} & 294.802 & 64.002 \\
      &                       & \cmark & \q{5514.820} & \textbf{\q{113.856}} & \textbf{\q{34.212}} \\
      & \multirow{2}{*}{GPTQ} & \xmark & 44807.926 & 106.715 & 27.839 \\
      &                       & \cmark & \q{N/A} & \textbf{\q{81.117}} & \textbf{\q{27.469}} \\
      & \multirow{2}{*}{AWQ}  & \xmark & 130.308 & 121.698 & 26.887 \\
      &                       & \cmark & \textbf{\q{81.606}} & \textbf{\q{93.260}} & \textbf{\q{25.592}} \\
    \midrule
    \multirow{6}{*}{INT2g32}
      & \multirow{2}{*}{RTN}  & \xmark & 20280.412 & 262.244 & 63.428 \\
      &                       & \cmark & \textbf{\q{1685.683}} & \textbf{\q{96.913}} & \textbf{\q{36.677}} \\
      & \multirow{2}{*}{GPTQ} & \xmark & 18292.635 & 152.169 & \textbf{29.163} \\
      &                       & \cmark & \q{N/A} & \textbf{\q{110.507}} & \q{30.465} \\
      & \multirow{2}{*}{AWQ}  & \xmark & 47850.137 & 60977.195 & 48520.398 \\
      &                       & \cmark & \textbf{\q{3741.642}} & \textbf{\q{47591.414}} & \textbf{\q{20185.246}} \\
    \midrule
    \multirow{6}{*}{INT2g64}
      & \multirow{2}{*}{RTN}  & \xmark & 9252.538 & 551.510 & 153.528 \\
      &                       & \cmark & \textbf{\q{1096.720}} & \textbf{\q{158.306}} & \textbf{\q{42.991}} \\
      & \multirow{2}{*}{GPTQ} & \xmark & N/A & 275.949 & \textbf{37.024} \\
      &                       & \cmark & \q{N/A} & \textbf{\q{187.477}} & \q{37.384} \\
      & \multirow{2}{*}{AWQ}  & \xmark & \textbf{202939.484} & \textbf{113584.867} & \textbf{79866.031} \\
      &                       & \cmark & \q{220728.234} & \q{117658.867} & \q{82598.511} \\
    \midrule
    \multirow{6}{*}{INT2g128}
      & \multirow{2}{*}{RTN}  & \xmark & 9685.755 & 1213.282 & 767.896 \\
      &                       & \cmark & \textbf{\q{4462.478}} & \textbf{\q{207.651}} & \textbf{\q{63.806}} \\
      & \multirow{2}{*}{GPTQ} & \xmark & 10694.694 & 395.689 & 56.685 \\
      &                       & \cmark & \q{N/A} & \textbf{\q{325.407}} & \textbf{\q{45.569}} \\
      & \multirow{2}{*}{AWQ}  & \xmark & \textbf{202164.484} & \textbf{113784.242} & \textbf{80543.727} \\
      &                       & \cmark & \q{222388.375} & \q{117059.742} & \q{82493.251} \\
    \midrule
    \multirow{6}{*}{INT2}
      & \multirow{2}{*}{RTN}  & \xmark & 31824.279 & \textbf{42619.883} & \textbf{26063.672} \\
      &                       & \cmark & \textbf{\q{10824.680}} & \q{55286.305} & \q{26077.172} \\
      & \multirow{2}{*}{GPTQ} & \xmark & N/A & 3868.426 & 2438.034 \\
      &                       & \cmark & \q{N/A} & \textbf{\q{3850.578}} & \textbf{\q{4050.844}} \\
      & \multirow{2}{*}{AWQ}  & \xmark & \textbf{183984.766} & 87673.695 & \textbf{90442.352} \\
      &                       & \cmark & \q{198744.750} & \textbf{\q{62160.063}} & \q{91939.883} \\
    \bottomrule
  \end{tabular}
\end{table}

\begin{table}[tb]
  \centering
  \small
  \caption{Perplexities ($\downarrow$) on C4 for Llama-2 (7B, 13B, 70B) under eight quantization settings.}
  \label{tab:gqep_c4_full}
  \setlength{\tabcolsep}{4pt}
  \renewcommand{\arraystretch}{1.12}

  \begin{tabular}{c|cc|ccc}
    \toprule
    \textbf{Bits} & \textbf{Method} & \textbf{QEP} &
    \textbf{Llama-2-7B} & \textbf{Llama-2-13B} & \textbf{Llama-2-70B} \\
    \midrule
    \multirow{6}{*}{INT4g128}
      & \multirow{2}{*}{RTN}  & \xmark & 7.584 & 6.869 & 5.826 \\
      &                       & \cmark &\textbf{\q{7.513}} & \textbf{\q{6.839}} & \textbf{\q{5.786}} \\ 
      & \multirow{2}{*}{GPTQ} & \xmark & 7.522 & 6.860 & 5.778 \\
      &                       & \cmark & \textbf{\q{7.421}} & \textbf{\q{6.828}} & \textbf{\q{5.770}} \\
      & \multirow{2}{*}{AWQ}  & \xmark & 7.443 & 6.840 & 5.772 \\
      &                       & \cmark & \textbf{\q{7.416}} & \textbf{\q{6.829}} & \textbf{\q{5.767}} \\
    \midrule
    \multirow{6}{*}{INT4}
      & \multirow{2}{*}{RTN}  & \xmark & 8.165 & 7.146 & 6.012 \\
      &                       & \cmark & \textbf{\q{7.945}} & \textbf{\q{7.067}} & \textbf{\q{5.947}} \\
      & \multirow{2}{*}{GPTQ} & \xmark & 7.866 & 7.069 & 5.905 \\
      &                       & \cmark & \textbf{\q{7.719}} & \textbf{\q{6.998}} & \textbf{\q{5.880}} \\
      & \multirow{2}{*}{AWQ}  & \xmark & 7.721 & 6.962 & 5.842 \\
      &                       & \cmark & \textbf{\q{7.634}} & \textbf{\q{6.932}} & \textbf{\q{5.828}} \\
    \midrule
    \multirow{6}{*}{INT3g128}
      & \multirow{2}{*}{RTN}  & \xmark & 8.977 & 7.582 & 6.266 \\
      &                       & \cmark & \textbf{\q{8.510}} & \textbf{\q{7.402}} & \textbf{\q{6.150}} \\
      & \multirow{2}{*}{GPTQ} & \xmark & 8.502 & 7.463 & 6.105 \\
      &                       & \cmark & \textbf{\q{8.185}} & \textbf{\q{7.316}} & \textbf{\q{6.072}} \\
      & \multirow{2}{*}{AWQ}  & \xmark & 8.300 & 7.310 & 6.036 \\
      &                       & \cmark & \textbf{\q{8.105}} & \textbf{\q{7.264}} & \textbf{\q{6.019}} \\
    \midrule
    \multirow{6}{*}{INT3}
      & \multirow{2}{*}{RTN}  & \xmark & 524.279 & 13.883 & 10.886 \\
      &                       & \cmark & \textbf{\q{21.436}} & \textbf{\q{10.284}} & \textbf{\q{8.202}} \\
      & \multirow{2}{*}{GPTQ} & \xmark & 11.780 & 8.826 & 7.067 \\
      &                       & \cmark & \textbf{\q{9.950}} & \textbf{\q{8.429}} & \textbf{\q{6.869}} \\
      & \multirow{2}{*}{AWQ}  & \xmark & 17.418 & 9.049 & 6.631 \\
      &                       & \cmark & \textbf{\q{13.934}} & \textbf{\q{8.257}} & \textbf{\q{6.353}} \\
    \midrule
    \multirow{6}{*}{INT2g32}
      & \multirow{2}{*}{RTN}  & \xmark & 225.440 & 13.879 & 9.720 \\
      &                       & \cmark & \textbf{\q{16.148}} & \textbf{\q{10.561}} & \textbf{\q{8.459}} \\
      & \multirow{2}{*}{GPTQ} & \xmark & 14.365 & 10.719 & 7.932 \\
      &                       & \cmark & \textbf{\q{11.839}} & \textbf{\q{9.685}} & \textbf{\q{7.717}} \\
      & \multirow{2}{*}{AWQ}  & \xmark & 9028.133 & 76591.883 & 57596.215 \\
      &                       & \cmark & \textbf{\q{51.811}} & \textbf{\q{49645.738}} & \textbf{\q{33026.816}} \\
    \midrule
    \multirow{6}{*}{INT2g64}
      & \multirow{2}{*}{RTN}  & \xmark & 553.766 & 30.445 & 15.155 \\
      &                       & \cmark & \textbf{\q{22.089}} & \textbf{\q{12.762}} & \textbf{\q{9.850}} \\
      & \multirow{2}{*}{GPTQ} & \xmark & 20.860 & 13.394 & 8.981 \\
      &                       & \cmark & \textbf{\q{14.084}} & \textbf{\q{11.039}} & \textbf{\q{8.508}} \\
      & \multirow{2}{*}{AWQ}  & \xmark & \textbf{164477.422} & \textbf{95241.625} & \textbf{64913.477} \\
      &                       & \cmark & \q{181582.719} & \q{98917.820} & \q{67203.359} \\
    \midrule
    \multirow{6}{*}{INT2g128}
      & \multirow{2}{*}{RTN}  & \xmark & 4811.772 & 131.665 & 47.878 \\
      &                       & \cmark & \textbf{\q{34.022}} & \textbf{\q{15.398}} & \textbf{\q{12.081}} \\
      & \multirow{2}{*}{GPTQ} & \xmark & 33.370 & 18.008 & 10.535 \\
      &                       & \cmark & \textbf{\q{18.184}} & \textbf{\q{12.704}} & \textbf{\q{9.433}} \\
      & \multirow{2}{*}{AWQ}  & \xmark & \textbf{168465.266} & \textbf{95617.305} & \textbf{65646.594} \\
      &                       & \cmark & \q{187329.625} & \q{98457.031} & \q{67248.492} \\
    \midrule
    \multirow{6}{*}{INT2}
      & \multirow{2}{*}{RTN}  & \xmark & \textbf{28258.385} & \textbf{52642.387} & \textbf{24912.074} \\
      &                       & \cmark & \q{108424.680} & \q{71050.250} & \q{29042.623} \\
      & \multirow{2}{*}{GPTQ} & \xmark & 3048.671 & \textbf{299.684} & 56.719 \\
      &                       & \cmark & \textbf{\q{276.638}} & \q{629.527} & \textbf{\q{30.874}} \\
      & \multirow{2}{*}{AWQ}  & \xmark & \textbf{156266.797} & 81233.602 & \textbf{73251.945} \\
      &                       & \cmark & \q{177576.750} & \textbf{\q{64098.504}} & \q{75607.211} \\
    \bottomrule
  \end{tabular}
\end{table}

\subsection{Detailed Accuracy Results for Individual Tasks}\label{app-subsec:detailed-acc-individual-task}
Due to space limitations, the main text reports only the average accuracy across three tasks. Here, we provide task-specific accuracies for PIQA (Table~\ref{tab:gqep_piqa_full}), StoryCloze (Table~\ref{tab:gqep_storycloze_full}), and ARC-Easy (Table~\ref{tab:gqep_arc_easy_full}), further confirming that QEP consistently improves layer-wise PTQ.

\begin{table}[tb]
  \centering
  \small
  \caption{Accuracy ($\uparrow$) on PIQA for Llama-2 (7B, 13B, 70B) under eight quantization settings.}
  \label{tab:gqep_piqa_full}
  \setlength{\tabcolsep}{4pt}
  \renewcommand{\arraystretch}{1.12}
  \begin{tabular}{c|cc|ccc}
    \toprule
    \textbf{Bits} & \textbf{Method} & \textbf{QEP} &
    \textbf{Llama-2-7B} & \textbf{Llama-2-13B} & \textbf{Llama-2-70B} \\
    \midrule
    \multirow{6}{*}{INT4g128}
      & \multirow{2}{*}{RTN}  & \xmark & \textbf{0.773} & \textbf{0.792} & 0.804 \\
      &                       & \cmark & \textbf{\q{0.773}} & \q{0.790} & \textbf{\q{0.806}} \\
      & \multirow{2}{*}{GPTQ} & \xmark & 0.770 & 0.789 & \textbf{0.807} \\
      &                       & \cmark & \textbf{\q{0.771}} & \textbf{\q{0.792}} & \q{0.806} \\
      & \multirow{2}{*}{AWQ}  & \xmark & \textbf{0.768} & 0.790 & 0.807 \\
      &                       & \cmark & \q{0.764} & \textbf{\q{0.791}} & \textbf{\q{0.810}} \\
    \midrule
    \multirow{6}{*}{INT4}
      & \multirow{2}{*}{RTN}  & \xmark & 0.763 & \textbf{0.789} & 0.811 \\
      &                       & \cmark & \textbf{\q{0.767}} & \q{0.788} & \textbf{\q{0.812}} \\
      & \multirow{2}{*}{GPTQ} & \xmark & 0.755 & \textbf{0.789} & 0.804 \\
      &                       & \cmark & \textbf{\q{0.761}} & \q{0.787} & \textbf{\q{0.811}} \\
      & \multirow{2}{*}{AWQ}  & \xmark & 0.760 & \textbf{0.789} & 0.807 \\
      &                       & \cmark & \textbf{\q{0.763}} & \q{0.784} & \textbf{\q{0.814}} \\
    \midrule
    \multirow{6}{*}{INT3g128}
      & \multirow{2}{*}{RTN}  & \xmark & 0.757 & 0.770 & 0.793 \\
      &                       & \cmark & \textbf{\q{0.761}} & \textbf{\q{0.779}} & \textbf{\q{0.806}} \\
      & \multirow{2}{*}{GPTQ} & \xmark & 0.758 & 0.778 & 0.806 \\
      &                       & \cmark & \textbf{\q{0.764}} & \textbf{\q{0.782}} & \textbf{\q{0.807}} \\
      & \multirow{2}{*}{AWQ}  & \xmark & 0.760 & \textbf{0.780} & \textbf{0.805} \\
      &                       & \cmark & \textbf{\q{0.765}} & \textbf{\q{0.780}} & \textbf{\q{0.805}} \\
    \midrule
    \multirow{6}{*}{INT3}
      & \multirow{2}{*}{RTN}  & \xmark & 0.563 & 0.705 & 0.724 \\
      &                       & \cmark & \textbf{\q{0.677}} & \textbf{\q{0.752}} & \textbf{\q{0.764}} \\
      & \multirow{2}{*}{GPTQ} & \xmark & 0.720 & 0.757 & 0.783 \\
      &                       & \cmark & \textbf{\q{0.745}} & \textbf{\q{0.770}} & \textbf{\q{0.791}} \\
      & \multirow{2}{*}{AWQ}  & \xmark & 0.647 & 0.760 & 0.787 \\
      &                       & \cmark & \textbf{\q{0.725}} & \textbf{\q{0.770}} & \textbf{\q{0.801}} \\
    \midrule
    \multirow{6}{*}{INT2g32}
      & \multirow{2}{*}{RTN}  & \xmark & 0.588 & 0.696 & 0.760 \\
      &                       & \cmark & \textbf{\q{0.693}} & \textbf{\q{0.735}} & \textbf{\q{0.771}} \\
      & \multirow{2}{*}{GPTQ} & \xmark & 0.690 & 0.732 & 0.772 \\
      &                       & \cmark & \textbf{\q{0.714}} & \textbf{\q{0.748}} & \textbf{\q{0.776}} \\
      & \multirow{2}{*}{AWQ}  & \xmark & 0.568 & 0.505 & \textbf{0.503} \\
      &                       & \cmark & \textbf{\q{0.702}} & \textbf{\q{0.514}} & \q{0.501} \\
    \midrule
    \multirow{6}{*}{INT2g64}
      & \multirow{2}{*}{RTN}  & \xmark & 0.597 & 0.614 & 0.714 \\
      &                       & \cmark & \textbf{\q{0.676}} & \textbf{\q{0.710}} & \textbf{\q{0.748}} \\ 
      & \multirow{2}{*}{GPTQ} & \xmark & 0.647 & 0.705 & 0.745 \\
      &                       & \cmark & \textbf{\q{0.677}} & \textbf{\q{0.713}} & \textbf{\q{0.765}} \\
      & \multirow{2}{*}{AWQ}  & \xmark & 0.502 & \textbf{0.506} & 0.502 \\
      &                       & \cmark & \textbf{\q{0.702}} & \textbf{\q{0.506}} & \textbf{\q{0.504}} \\
    \midrule
    \multirow{6}{*}{INT2g128}
      & \multirow{2}{*}{RTN}  & \xmark & 0.511 & 0.566 & 0.635 \\
      &                       & \cmark & \textbf{\q{0.652}} & \textbf{\q{0.678}} & \textbf{\q{0.721}} \\
      & \multirow{2}{*}{GPTQ} & \xmark & 0.581 & 0.639 & 0.715 \\
      &                       & \cmark & \textbf{\q{0.659}} & \textbf{\q{0.683}} & \textbf{\q{0.747}} \\
      & \multirow{2}{*}{AWQ}  & \xmark & \textbf{0.501} & 0.505 & \textbf{0.503} \\
      &                       & \cmark & \textbf{\q{0.501}} & \textbf{\q{0.507}} & \textbf{\q{0.503}} \\
    \midrule
    \multirow{6}{*}{INT2}
      & \multirow{2}{*}{RTN}  & \xmark & 0.509 & 0.493 & 0.499 \\
      &                       & \cmark & \textbf{\q{0.510}} & \textbf{\q{0.506}} & \q{0.510} \\
      & \multirow{2}{*}{GPTQ} & \xmark & \textbf{0.500} & \textbf{0.509} & 0.511 \\
      &                       & \cmark & \q{0.493} & \textbf{\q{0.507}} & \textbf{\q{0.544}} \\
      & \multirow{2}{*}{AWQ}  & \xmark & \textbf{0.507} & 0.504 & 0.502 \\
      &                       & \cmark & \q{0.505} & \q{0.504} & \q{0.504} \\
    \bottomrule
  \end{tabular}
\end{table}

\begin{table}[tb]
  \centering
  \small
  \caption{Accuracy ($\uparrow$) on StoryCloze for Llama-2 (7B, 13B, 70B) under eight quantization settings.}
  \label{tab:gqep_storycloze_full}
  \setlength{\tabcolsep}{4pt}
  \renewcommand{\arraystretch}{1.12}

  \begin{tabular}{c|cc|ccc}
    \toprule
    \textbf{Bits} & \textbf{Method} & \textbf{QEP} &
    \textbf{Llama-2-7B} & \textbf{Llama-2-13B} & \textbf{Llama-2-70B} \\
    \midrule
    \multirow{6}{*}{INT4g128}
      & \multirow{2}{*}{RTN}  & \xmark & 0.765 & 0.785 & 0.791 \\
      &                       & \cmark & \textbf{\q{0.770}} & \textbf{\q{0.788}} & \textbf{\q{0.794}} \\
      & \multirow{2}{*}{GPTQ} & \xmark & 0.768 & 0.784 & 0.793 \\
      &                       & \cmark & \textbf{\q{0.771}} & \textbf{\q{0.789}} & \textbf{\q{0.798}} \\
      & \multirow{2}{*}{AWQ}  & \xmark & \textbf{0.777} & 0.782 & 0.792 \\
      &                       & \cmark & \textbf{\q{0.777}} & \textbf{\q{0.785}} & \textbf{\q{0.798}} \\
    \midrule
    \multirow{6}{*}{INT4}
      & \multirow{2}{*}{RTN}  & \xmark & 0.756 & \textbf{0.777} & 0.796 \\
      &                       & \cmark & \textbf{\q{0.763}} & \textbf{\q{0.777}} & \textbf{\q{0.798}} \\
      & \multirow{2}{*}{GPTQ} & \xmark & 0.765 & \textbf{0.776} & \textbf{0.794} \\
      &                       & \cmark & \textbf{\q{0.766}} & \q{0.775} & \q{0.792} \\
      & \multirow{2}{*}{AWQ}  & \xmark & 0.760 & 0.774 & 0.789 \\
      &                       & \cmark & \textbf{\q{0.766}} & \textbf{\q{0.777}} & \textbf{\q{0.794}} \\
    \midrule
    \multirow{6}{*}{INT3g128}
      & \multirow{2}{*}{RTN}  & \xmark & 0.749 & 0.766 & \textbf{0.790} \\
      &                       & \cmark & \textbf{\q{0.756}} & \textbf{\q{0.773}} & \q{0.789} \\
      & \multirow{2}{*}{GPTQ} & \xmark & \textbf{0.763} & \textbf{0.776} & 0.793 \\
      &                       & \cmark & \q{0.759} & \q{0.770} & \textbf{\q{0.796}} \\
      & \multirow{2}{*}{AWQ}  & \xmark & \textbf{0.761} & 0.767 & \textbf{0.795} \\
      &                       & \cmark & \textbf{\q{0.761}} & \textbf{\q{0.782}} & \textbf{\q{0.795}} \\
    \midrule
    \multirow{6}{*}{INT3}
      & \multirow{2}{*}{RTN}  & \xmark & 0.546 & 0.669 & 0.738 \\
      &                       & \cmark & \textbf{\q{0.672}} &\textbf{ \q{0.728}} & \textbf{\q{0.776}} \\
      & \multirow{2}{*}{GPTQ} & \xmark & 0.722 & 0.752 & 0.780 \\
      &                       & \cmark & \textbf{\q{0.745}} & \textbf{\q{0.766}} & \textbf{\q{0.782}} \\
      & \multirow{2}{*}{AWQ}  & \xmark & 0.689 & \textbf{0.767} & \textbf{0.787} \\
      &                       & \cmark & \textbf{\q{0.702}} & \q{0.764} & \q{0.782} \\
    \midrule
    \multirow{6}{*}{INT2g32}
      & \multirow{2}{*}{RTN}  & \xmark & 0.645 & 0.668 & 0.745 \\
      &                       & \cmark & \textbf{\q{0.704}} & \textbf{\q{0.721}} & \textbf{\q{0.776}} \\
      & \multirow{2}{*}{GPTQ} & \xmark & 0.758 & 0.715 & 0.724 \\
      &                       & \cmark & \textbf{\q{0.763}} & \textbf{\q{0.748}} & \textbf{\q{0.766}} \\
      & \multirow{2}{*}{AWQ}  & \xmark & 0.660 & 0.511 & 0.516 \\
      &                       & \cmark & \textbf{\q{0.703}} & \textbf{\q{0.570}} & \textbf{\q{0.569}} \\
    \midrule
    \multirow{6}{*}{INT2g64}
      & \multirow{2}{*}{RTN}  & \xmark & 0.607 & 0.617 & 0.718 \\
      &                       & \cmark & \textbf{\q{0.670}} & \textbf{\q{0.696}} & \textbf{\q{0.766}} \\
      & \multirow{2}{*}{GPTQ} & \xmark & 0.654 & 0.686 & 0.756 \\
      &                       & \cmark & \textbf{\q{0.712}} & \textbf{\q{0.720}} & \textbf{\q{0.758}} \\
      & \multirow{2}{*}{AWQ}  & \xmark & \textbf{0.476} & \textbf{0.479} & \textbf{0.476} \\
      &                       & \cmark & \q{0.474} & \textbf{\q{0.479}} & \q{0.475} \\
    \midrule
    \multirow{6}{*}{INT2g128}
      & \multirow{2}{*}{RTN}  & \xmark & 0.509 & 0.577 & 0.647 \\
      &                       & \cmark & \textbf{\q{0.651}} & \textbf{\q{0.677}} & \textbf{\q{0.741}} \\
      & \multirow{2}{*}{GPTQ} & \xmark & 0.588 & 0.634 & 0.724 \\
      &                       & \cmark & \textbf{\q{0.649}} & \textbf{\q{0.690}} & \textbf{\q{0.753}} \\
      & \multirow{2}{*}{AWQ}  & \xmark & \textbf{0.475} & \textbf{0.478} & \textbf{0.476} \\
      &                       & \cmark & \textbf{\q{0.475}} & \textbf{\q{0.478}} & \textbf{\q{0.476}} \\
    \midrule
    \multirow{6}{*}{INT2}
      & \multirow{2}{*}{RTN}  & \xmark & 0.468 & \textbf{0.491} & \textbf{0.482} \\
      &                       & \cmark & \textbf{\q{0.488}} & \q{0.487} & \textbf{\q{0.482}} \\
      & \multirow{2}{*}{GPTQ} & \xmark & 0.485 & 0.501 & 0.539 \\
      &                       & \cmark & \textbf{\q{0.514}} & \textbf{\q{0.513}} & \textbf{\q{0.589}} \\
      & \multirow{2}{*}{AWQ}  & \xmark & \textbf{0.489} & \textbf{0.478} & 0.475 \\
      &                       & \cmark & \q{0.482} & \q{0.476} & \textbf{\q{0.477}} \\
    \bottomrule
  \end{tabular}
\end{table}

\begin{table}[tb]
  \centering
  \small
  \caption{Accuracy ($\uparrow$) on ARC-Easy for Llama-2 (7B, 13B, 70B) under eight quantization settings.}
  \label{tab:gqep_arc_easy_full}
  \setlength{\tabcolsep}{4pt}
  \renewcommand{\arraystretch}{1.12}

  \begin{tabular}{c|cc|ccc}
    \toprule
    \textbf{Bits} & \textbf{Method} & \textbf{QEP} &
    \textbf{Llama-2-7B} & \textbf{Llama-2-13B} & \textbf{Llama-2-70B} \\
    \midrule
    \multirow{6}{*}{INT4g128}
      & \multirow{2}{*}{RTN}  & \xmark & \textbf{0.554} & 0.567 & \textbf{0.596} \\
      &                       & \cmark & \q{0.540} & \textbf{\q{0.572}} & \textbf{\q{0.596}} \\
      & \multirow{2}{*}{GPTQ} & \xmark & \textbf{0.531} & 0.573 & 0.586 \\
      &                       & \cmark & \q{0.521} & \textbf{\q{0.579}} & \textbf{\q{0.592}} \\
      & \multirow{2}{*}{AWQ}  & \xmark & \textbf{0.537} & 0.577 & 0.585 \\
      &                       & \cmark & \q{0.526} & \textbf{\q{0.580}} & \textbf{\q{0.592}} \\
    \midrule
    \multirow{6}{*}{INT4}
      & \multirow{2}{*}{RTN}  & \xmark & 0.521 & \textbf{0.582} & 0.590 \\
      &                       & \cmark & \textbf{\q{0.524}} & \q{0.574} & \textbf{\q{0.593}} \\
      & \multirow{2}{*}{GPTQ} & \xmark & \textbf{0.525} & \textbf{0.575} & \textbf{0.594} \\
      &                       & \cmark & \q{0.512} & \q{0.570} & \q{0.589} \\
      & \multirow{2}{*}{AWQ}  & \xmark & 0.529 & 0.572 & 0.580 \\
      &                       & \cmark & \textbf{\q{0.532}} & \textbf{\q{0.577}} & \textbf{\q{0.591}} \\
    \midrule
    \multirow{6}{*}{INT3g128}
      & \multirow{2}{*}{RTN}  & \xmark & \textbf{0.528} & \textbf{0.569} & \textbf{0.575} \\
      &                       & \cmark & \q{0.517} & \q{0.556} & \q{0.572} \\
      & \multirow{2}{*}{GPTQ} & \xmark & 0.521 & \textbf{0.568} & \textbf{0.580} \\
      &                       & \cmark & \textbf{\q{0.515}} & \textbf{\q{0.568}} & \q{0.569} \\
      & \multirow{2}{*}{AWQ}  & \xmark & \textbf{0.534} & \textbf{0.561} & \textbf{0.597} \\
      &                       & \cmark & \q{0.527} & \textbf{\q{0.561}} & \q{0.592} \\
    \midrule
    \multirow{6}{*}{INT3}
      & \multirow{2}{*}{RTN}  & \xmark & 0.322 & 0.450 & \textbf{0.459} \\
      &                       & \cmark & \textbf{\q{0.391}} & \textbf{\q{0.485}} & \q{0.541} \\
      & \multirow{2}{*}{GPTQ} & \xmark & 0.468 & 0.514 & 0.550 \\
      &                       & \cmark & \textbf{\q{0.474}} & \textbf{\q{0.520}} & \textbf{\q{0.551}} \\
      & \multirow{2}{*}{AWQ}  & \xmark & 0.416 & 0.539 & 0.588 \\
      &                       & \cmark & \textbf{\q{0.452}} & \textbf{\q{0.540}} & \textbf{\q{0.602}} \\
    \midrule
    \multirow{6}{*}{INT2g32}
      & \multirow{2}{*}{RTN}  & \xmark & 0.339 & 0.445 & 0.533 \\
      &                       & \cmark & \textbf{\q{0.426}} & \textbf{\q{0.474}} & \textbf{\q{0.557}} \\
      & \multirow{2}{*}{GPTQ} & \xmark & 0.421 & 0.481 & 0.506 \\
      &                       & \cmark & \textbf{\q{0.441}} & \textbf{\q{0.486}} & \textbf{\q{0.547}} \\
      & \multirow{2}{*}{AWQ}  & \xmark & 0.352 & 0.272 & \textbf{0.263} \\
      &                       & \cmark & \textbf{\q{0.449}} & \textbf{\q{0.280}} & \textbf{\q{0.263}} \\
    \midrule
    \multirow{6}{*}{INT2g64}
      & \multirow{2}{*}{RTN}  & \xmark & 0.332 & 0.371 & 0.467 \\
      &                       & \cmark & \textbf{\q{0.390}} & \textbf{\q{0.430}} & \textbf{\q{0.557}} \\
      & \multirow{2}{*}{GPTQ} & \xmark & 0.377 & 0.455 & 0.485 \\
      &                       & \cmark & \textbf{\q{0.404}} & \textbf{\q{0.458}} & \textbf{\q{0.548}} \\
      & \multirow{2}{*}{AWQ}  & \xmark & \textbf{0.266} & \textbf{0.270} & 0.262 \\
      &                       & \cmark & \q{0.265} & \textbf{\q{0.270}} & \textbf{\q{0.263}} \\
    \midrule
    \multirow{6}{*}{INT2g128}
      & \multirow{2}{*}{RTN}  & \xmark & 0.269 & 0.253 & 0.395 \\
      &                       & \cmark & \textbf{\q{0.376}} & \textbf{\q{0.407}} & \textbf{\q{0.479}} \\
      & \multirow{2}{*}{GPTQ} & \xmark & 0.338 & 0.383 & 0.443 \\
      &                       & \cmark & \textbf{\q{0.367}} & \textbf{\q{0.418}} & \textbf{\q{0.508}} \\
      & \multirow{2}{*}{AWQ}  & \xmark & \textbf{0.266} & \textbf{0.269} & 0.260 \\
      &                       & \cmark & \q{0.265} & \textbf{\q{0.269}} & \textbf{\q{0.261}} \\
    \midrule
    \multirow{6}{*}{INT2}
      & \multirow{2}{*}{RTN}  & \xmark & \textbf{0.265} & 0.253 & \textbf{0.263} \\
      &                       & \cmark & \q{0.262} & \textbf{\q{0.264}} & \q{0.261} \\
      & \multirow{2}{*}{GPTQ} & \xmark & 0.263 & \textbf{0.256} & 0.257 \\
      &                       & \cmark & \textbf{\q{0.272}} & \q{0.265} & \textbf{\q{0.281}} \\
      & \multirow{2}{*}{AWQ}  & \xmark &\textbf{0.267} & \textbf{0.270} & \textbf{0.262} \\
      &                       & \cmark & \q{0.262} & \textbf{\q{0.270}} & \q{0.261} \\
    \bottomrule
  \end{tabular}
\end{table}

\subsection{Stability of QuIP Results Across Random Seeds}\label{app-subsec:stability-quip}
We assess the stability of QuIP-only experiments by averaging five independent runs per configuration. Model sizes, quantization levels, and benchmarks align with the main Experiments section.
Figure \ref{fig:quip-results} plots QuIP with or without QEP at three quantization levels. Each marker is the mean of five seeds, and the error bars show the standard error of the mean. The top row gives perplexity on WikiText 2; the bottom row reports mean normalized accuracy on ARC easy, PIQA, and StoryCloze.
Seed-to-seed variation is small and does not change the main conclusions. QEP-QuIP keeps its advantage, especially at INT3 and INT2.
The main text lists the best seed per configuration for consistency with past work. This appendix confirms that the gains are not seed-specific but robust and reproducible, supporting using QEP.

\begin{wraptable}{r}{0.46\textwidth}
\vspace{-40pt}
\centering
\small
\caption{WikiText-2 perplexity for LLaMA-2-7B at different bit-widths. NaN denotes divergence.}
\begin{tabular}{lccc}
\toprule
\textbf{Method} & \textbf{INT4} & \textbf{INT3} & \textbf{INT2} \\
\midrule
RTN+QEP & 6.017 & 17.309 & 97153.266 \\
GPTQ+QEP & 5.933 & 7.898 & 7214.328 \\
AWQ+QEP & 5.756 & 11.131 & 229888.406 \\
\textbf{QuIP+QEP} & \textbf{5.753} & \textbf{6.154} & \textbf{11.972} \\
OmniQuant & 5.880 & 7.065 & NaN \\
\bottomrule
\end{tabular}
\label{tab:omniquant}
\end{wraptable}

\subsection{Comparison with OmniQuant Baseline}\label{app-subsec:omniquant}

For completeness, we compare QEP-enhanced \emph{layer-wise} PTQ with \emph{block-wise} OmniQuant \citep{shao2023omniquant} on LLaMA-2-7B using WikiText-2 perplexity; lower values indicate better performance.
As shown in Table~\ref{tab:omniquant}, QuIP+QEP achieves the lowest perplexity at INT4/INT3 and remains stable at INT2, while OmniQuant diverges. 
These findings align with recent PTQ benchmarks indicating OmniQuant's underperformance relative to layer-wise PTQ~\citep{zhao2025benchmarking}.

\end{document}